\pdfoutput=1
\documentclass[11pt]{article}

\usepackage{acl}

\usepackage{times}
\usepackage{latexsym}
\usepackage{bm}
\usepackage{colortbl}

\usepackage[T1]{fontenc}
\usepackage[utf8]{inputenc}

\usepackage{microtype}

\usepackage{inconsolata}
\usepackage{amsmath}
\usepackage{mathtools}
\usepackage{graphicx}
\usepackage{centernot}
\usepackage{amsfonts}
\usepackage{booktabs}

\usepackage{enumitem}
\usepackage{soul}
\usepackage{lineno}
\usepackage{algorithm}
\usepackage{algorithmic}
\usepackage{multirow}
\usepackage{soul}
\usepackage{tikz}
\usepackage{titletoc}

\definecolor{maroon}{cmyk}{0,0.87,0.68,0.32}
\definecolor{customgreen}{cmyk}{39, 0, 39, 7}

\definecolor{navyblue}{cmyk}{1, 0.57, 0, 0.40}
\definecolor{darkorange}{cmyk}{0, 0.65, 0.87, 0.15}
\definecolor{purple}{cmyk}{0.45, 0.86, 0, 0.12}
\definecolor{teal}{cmyk}{0.85, 0.21, 0.30, 0.01}
\definecolor{olive}{cmyk}{0.44, 0, 0.95, 0.45}
\definecolor{darkred}{cmyk}{0, 1, 1, 0.5}
\definecolor{darkcyan}{cmyk}{1, 0, 0, 0.5}
\definecolor{darkmagenta}{cmyk}{0, 1, 0, 0.5}
\newcommand\scenariocount{$37$\ }

\title{
Towards Quantifying Commonsense Reasoning with Mechanistic Insights
}

\author{
  Abhinav Joshi \qquad Areeb Ahmad \qquad Divyaksh Shukla \qquad Ashutosh Modi \\
  Department of Computer Science and Engineering\\
  Indian Institute of Technology Kanpur (IIT Kanpur)\\
  \texttt{\{ajoshi,areeb,divyaksh,ashutoshm\}@cse.iitk.ac.in} \\
} 

\begin{document}
\maketitle

\begin{abstract}
Commonsense reasoning deals with the implicit knowledge that is well understood by humans and typically acquired via interactions with the world. In recent times, commonsense reasoning and understanding of various LLMs have been evaluated using text-based tasks. In this work, we argue that a proxy of this understanding can be maintained as a graphical structure that can further help to perform a rigorous evaluation of commonsense reasoning abilities about various real-world activities. We create an annotation scheme for capturing this implicit knowledge in the form of a graphical structure for 37 daily human activities. We find that the created resource can be used to frame an enormous number of commonsense queries ($\sim 10^{17}$), facilitating rigorous evaluation of commonsense reasoning in LLMs. Moreover, recently, the remarkable performance of LLMs has raised questions about whether these models are truly capable of reasoning in the wild and, in general, how reasoning occurs inside these models. In this resource paper, we bridge this gap by proposing design mechanisms that facilitate research in a similar direction. Our findings suggest that the reasoning components are localized in LLMs that play a prominent role in decision-making when prompted with a commonsense query.
% } 
% \AMC{maybe we should write 1-2 lines about main finding in here in slightly ``perhaps" tone}
% \st{More recently, LLMs have shown remarkable performance in various commonsense reasoning tasks, raising questions about whether these models are truly capable of reasoning in the wild and how reasoning occurs inside these models. 
% Moreover, tools to tease apart the decision-making happening inside these models remain limited.
% In this work, we bridge this gap by proposing design mechanisms that facilitate research in a similar direction.}  
\end{abstract}

\section{Introduction} \label{sec:intro}
%\vspace{-2mm}
% 
% Though there has been an immense amount of research in validating the commonsense reasoning of machine learning algorithms, a rigorous evaluation is missing from the literature. In this work, we aim to bridge this gap by proposing a scheme to create an enormous number of commonsense reasoning questions about simple real-world activities that are well understood by humans.

\begin{figure}[t]
\centering
 \includegraphics[width=\linewidth]{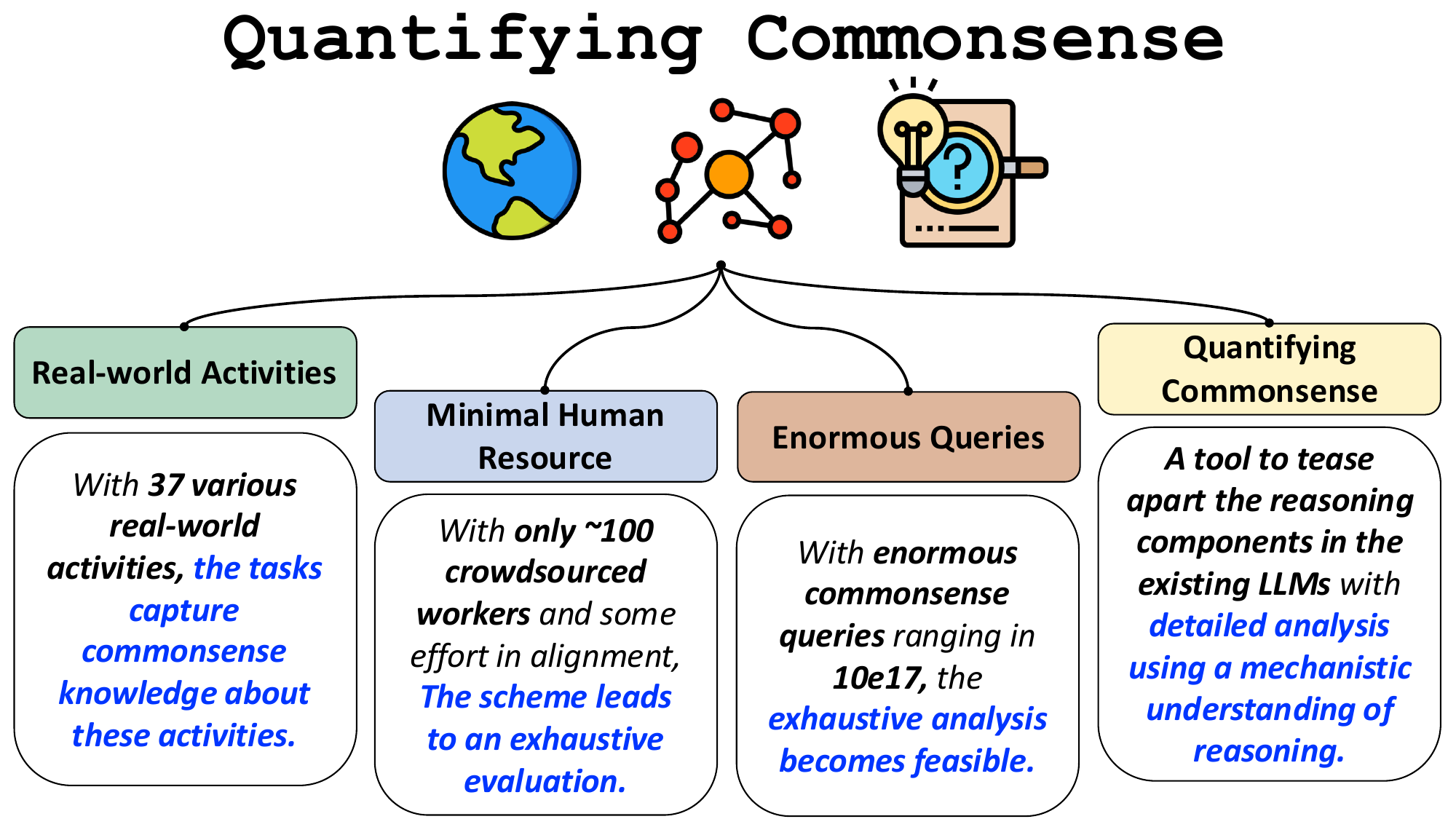}
  \caption{Quantifying commonsense reasoning in Large Langauge Models (LLMs). 
  % \AMC{in the last block, we should not use the word ``benchmarking" as later we specifically mention that our aim is not to do so, also we should include a point about understanding the reasoning mechanisms in LLMs.}
  }
  \label{fig:commonsense-thumbnail-intro}
  % \vspace{-6mm}
\end{figure}

\begin{figure*}[t]
\begin{center}
\includegraphics[width=0.85\linewidth]{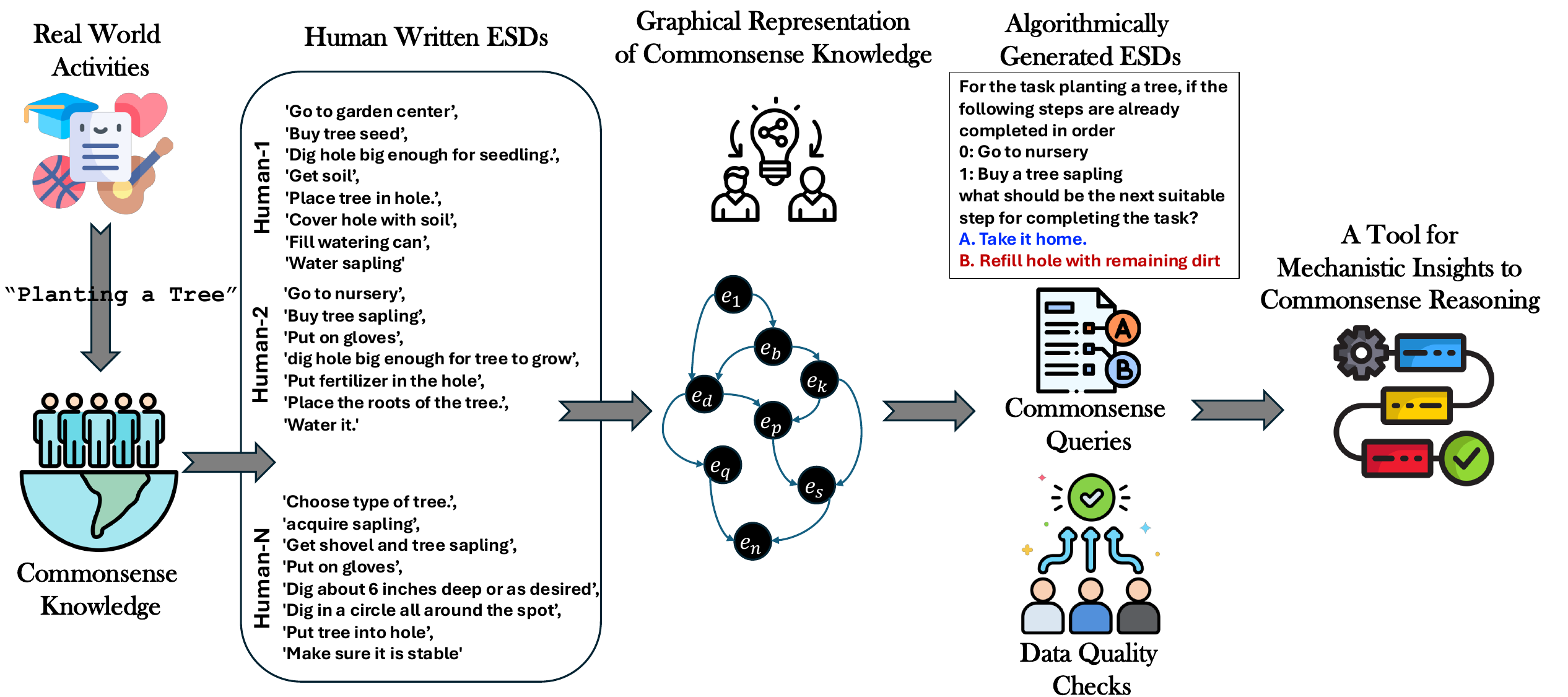} 
\caption{The figure provides an overview of the proposed resource. Real-world activities (well understood by humans) are considered to capture commonsense knowledge about these activities via human crowdsource workers. These ESDs are used to create a graphical representation of these activities and the underlying commonsense knowledge. The graphical representations help create enormous commonsense queries ($\sim 10^{17}$ queries per activity). The created resource of commonsense queries is reverified via data quality checks from humans. The overall flexibility attained using the graphical representations helps tease apart the reasoning mechanisms of LLMs, creating a tool for mechanistic insights into commonsense reasoning.}
% \label{fig:compact_graphs_changing_batteries}
\label{fig:quantifying_commonsense_thumbnail}
\end{center}
\vspace{-5mm}
\end{figure*}

The growth of Large Language Models (LLMs) performing well on a wide variety of commonsense reasoning benchmarks \cite{west-etal-2023-novacomet,bosselut-etal-2019-comet,hwang2021cometatomic,park2020visualcomet} raises the question of whether LLMs are truly capable of reasoning in a more practical setting of real-world daily activities that involve commonsense. Though in the past, a wide range of benchmarks/datasets (information sources) have been proposed, building a benchmark with exhaustive and rigorous analysis has always remained a challenge. To quantify the commonsense reasoning abilities of LLMs in an exhaustive manner, one would require a few primary features about an information resource 1) the information source should consider real-world tasks, well understood by humans (capturing commonsense) 2) the information resource should be exhaustive, containing all possible ways of performing a task, and, 3) the information resource should support creating reasoning questions, that help in 
% teasing apart the contained commonsense knowledge 
understanding of reasoning mechanisms of models
via marginalization with multiple samples. 
% \AMC{\st{We should also have a point on that it should enable understanding of reasoning mechanisms in LLMs.}} 
We found that 
\textit{``Scripts''} \citep{SCHANK1975237,schank1975scripts} help create a tangible framework 
% a concrete medium 
that satisfies all these requirements. 
Scripts are defined as a sequence of events describing a prototypical activity, such as `going to a restaurant,' `baking a cake,' etc., capturing commonsense knowledge about the world \citep{schank1975scripts,modi-etal-2016-inscript,wanzare-etal-2016-crowdsourced,ostermann-etal-2018-mcscript,modi-2016-event,Modi17-Thesis, modi-etal-2017-modeling,modi-titov-2014-inducing}. Since all the real-world tasks are generic, writing about steps/events while performing the activity can be done in an enormous number of different ways. Additionally, these activities are easy to reason about, and previous works \cite{modi2013learning,modi-titov-2014-inducing,modi-etal-2017-modeling} have used them to create commonsense reasoning queries, assessing the quality of acquired commonsense knowledge. 
Moreover, they inherently provide a structure that helps facilitate marginalization across different variations, opening new directions for localizing information  \cite{meng2023locatingeditingfactualassociations, mcgrath2023hydraeffectemergentselfrepair, wang2022interpretabilitywildcircuitindirect,goldowskydill2023localizingmodelbehaviorpath}\
% AJ{cite for localization} 
contained in the decision-making process of commonsense reasoning.
% (aka circuit discovery)
%For our rigorous analysis, we consider a crowdsourced resource DeScript \cite{wanzare-etal-2016-crowdsourced}, and build our framework over it to facilitate the rigorous evaluation of commonsense knowledge. 

\noindent In this work, we propose a generic scheme for rigorously evaluating commonsense knowledge and understanding of LLMs via commonsense reasoning questions. For a framework devised to validate the commonsense understanding of implicit commonsense knowledge, it becomes imperative to consider the dataset directly coming from humans (i.e. written and annotated by humans with minimal synthetic intervention). Hence, for our analysis, we consider a crowdsourced commonsense resource about daily human activities
called as DeScript \cite{wanzare-etal-2016-crowdsourced}. 
% \AJ{add a line about the sole purpose not being the benchmarking}
% \AM{
We create a directed graph from the DeScript corpus, which is subsequently used (via an algorithm) to generate commonsense reasoning questions about various activities. LLMs are then evaluated for commonsense reasoning via these questions.  Further, we investigate where does commonsense reside in the pretrained autoregressive transformer-based models. In particular, we use activation path patching to localize the decision-making for commonsense reasoning in these models. We find that the proposed framework provides promising flexibility for such analysis and will help facilitate future research in Mechanistic Interpretability for commonsense reasoning. We make the following contributions: 
\begin{itemize}[noitemsep,nosep,leftmargin=*] %\itemsep-0.3em 
    \item 
    % \AM{
    We provide resources for creating directed commonsense knowledge graph for \scenariocount scenarios (daily human activities). These graphical representations of human activities are suitable to act as a proxy for comprehending the underlying commonsense knowledge about these activities. Fig. \ref{fig:commonsense-thumbnail-intro} shows the key features of the framework.
    % } 
    % \st{We provide resources for aligning events in (scenariocount) scenarios (daily human activities) and provide the respective directed graphs to capture a rich understanding of these activities. These rich graphical representations of human activities are suitable to act as a proxy for comprehending the underlying commonsense knowledge about these activities. Fig  (fig:commonsense-thumbnail-intro) shows the key features of the proposed work.} 
    \item We propose a generic scheme (based on graphs) to create prompts that help validate the commonsense reasoning and language understanding.
    % of LLMs. 
    \item Via experimentation with 6 open-weight models  \texttt{gpt-neo-1.3B \cite{gpt-neo}}, \texttt{gpt-j-6B} \cite{gpt-j}, \texttt{phi-2} \cite{javaheripi2023phi}, \texttt{Llama2-7b} \cite{touvron2023llama2openfoundation}, \texttt{Mistral-7B} \cite{jiang2023mistral7b}, and \texttt{Llama-3-8B} \cite{grattafiori2024llama3herdmodels}, 
%     Phi \cite{javaheripi2023phi}
%     GPT-Neo 2.7B, from GPT-Neo family \cite{gpt-neo};
% GPT-J 6B \cite{gpt-j};
% Llama2 13B Chat, from Llama2 family  \cite{touvron2023llama2openfoundation};
% Llama3 8B, Llama3 8B-Instruct, from Llama3 family \cite{grattafiori2024llama3herdmodels}; and   
% Mistral 7B    \cite{jiang2023mistral7b}. 
    % \texttt{Llama2-7b, Mistral-7B, Llama-3-8B, gpt-neo-1.3B,gpt-j-6B, and phi-2}, we 
    % show an interesting trend about 
    we highlight trends and
    gaps in commonsense knowledge, understanding, and reasoning abilities of these models. %\AMC{Do we also mention about Claude model?}
    % \AJ{Overall, our findings suggest that commonsense reasoning is still challenging for the line of open-weight LLMs that we tested, and the ability to generate the correct sensible sequence may not be the correct measure to look at when considering the understanding of these activities by LLMs.} 
    % \AM{\st{Here we are making strong claim that is true for all LLMs but this may not be the case unless we have checked it on all including GPT-4, Clout, etc. Maybe we should tone down the claim.}}
    % \item \AM{Possibly we should include the interoperability aspects, this will exclude close models like GPT}
    \item 
    % \AM{
    As a use-case for the proposed dataset, with the aim of understanding the reasoning process,
    % } 
    we propose design mechanisms to tease apart the decision-making happening inside pretrained LLMs. 
    % In particular, 
    % We show that 
    The high flexibility of the proposed framework helps to provide a more decisive finding about commonsense reasoning happening inside these models.
    \item We perform localization experiments over \texttt{phi-2} (being both computationally moderate with better performance) and investigate the commonsense knowledge reasoning in detail. We release the dataset/code via GitHub: \url{https://github.com/Exploration-Lab/CoReMech}.
\end{itemize}

\section{Methodology}
%%\vspace{-2mm}

In this section, we provide details about the scheme created for a rigorous/exhaustive analysis of commonsense reasoning abilities related to daily real-world activities and how an enormous number of commonsense queries (ranging $\sim 10^{17}$ on average per scenario) can be created to evaluate the quality of commonsense understanding in LLMs. (Figure \ref{fig:quantifying_commonsense_thumbnail} provides an overview of the proposed scheme)%We start with the DeScript dataset description and provide details about the alignment annotations performed over the dataset. These alignment annotations help create a graphical representation of these stereotypical activities that are further used to create enormous reasoning questions about the activities. 

\noindent\textbf{Dataset:} As outlined earlier, we use 
% \AM{
crowdsourced resource DeScript \cite{wanzare-etal-2016-crowdsourced}, which provides a telegrammic-style version of script event sequences (referred to as Event Sequence Descriptions (ESDs)) for various stereotypical human activities. DeScript provides a list of $40$ stereotypical human activities (each referred to as a scenario or activity) along with $\sim100$ ESDs provided by crowd-sourced workers for each of the $40$ scenarios. DeScript also annotates $10$ scenarios by grouping similar events  (telegrammic steps). For example, in a scenario like \texttt{``Washing Dishes,''}, the events like \texttt{``dry utensils''} and \texttt{``clean utensils with a clean, dry cloth''} are grouped. In this work, we extend the annotations and provide the alignments for the remaining $30$ scenarios, leading to a rich resource of $37$ daily activity scenarios (3 scenarios are discarded as these were found to be too noisy). 
We create a directed graph with the help of aligned sequences (coming from annotations), consolidating information supplied by $\sim 100$ crowd workers into a single graph. Using the graph, we devise a scheme to generate commonsense reasoning questions about these activities. The complete list of the considered scenarios is presented in Table \ref{tab:graph_details_number_of_paths_degree}. 
% } \st{we use a crowdsourced DeScript dataset having telegrammic-style descriptions of various scenarios. } 
% \AMC{understanding will become easier if we have a small diagram similar to one we had in script world which shows ESDs, alignment, graph}

\begin{table}[t]
    \centering
    \tiny
    \renewcommand{\arraystretch}{1}
    \setlength\tabcolsep{2pt}
    \resizebox{\linewidth}{!}{%
    \begin{tabular}{cccc}
    \toprule
    Scenario/Activity                           & \begin{tabular}[c]{@{}c@{}} Deg. \end{tabular} & Total possible ESDs              \\
    \midrule
    
    \texttt{baking a \text{cake}} & 3.6 & $4.0e+{26}$ \\
    \texttt{borrowing book from \text{Library}} & 3.7 & $3.1e+{19}$ \\
    \texttt{changing batteries in \text{alarm clock}} & 5.8 & $8.1e+19$ \\
    \texttt{checking in an \text{airport}} & 8.6 & $7.7e+23$\\
    \texttt{cleaning up a \text{flat}} & 7.4 & $1.1e+20$\\
    \texttt{cooking \text{pasta}} & 5.4 & $1.1e+22$\\
    \texttt{doing \text{laundary}} & 9.5 & $5.0e+38$\\
    \texttt{eating in a fast \text{food restaurant}} & 6.7 & $6.9e+27$\\
    \texttt{flying in an \text{airplane}} & 3.6 & $2.6e+{30}$ \\
    \texttt{fueling a \text{car}} & 8.2 & $4.6e+29$\\
    \texttt{getting a \text{haircut}} & 3.7 & $4.0e+{28}$\\
    \texttt{going grocery \text{shopping}} & 3.7 & $2.3e+{26}$ \\
    \texttt{going on a \text{Train}} & 3.7 & $3.1e+{21}$ \\
    \texttt{going to the \text{dentist}} & 6.6 & $7.8e+23$\\
    \texttt{going to the \text{swimming pool}} & 7.2 & $1.5e+16$ \\
    \texttt{going to the \text{theatre}} & 6.3 & $8.1e+16$\\
    \texttt{going to the sauna} & 7.3 & $1.3e+22$\\
    % \texttt{going to a \text{fueneral} } & 4.0 & $2.6e+05$\\
    \texttt{going \text{bowling} } & 9.5 & $1.8e+37$\\
    \texttt{having a \text{barbeque}} & 6.8 & $6.5e+20$\\
    \texttt{ironing \text{Laundary}} & 7.8 & $2.1e+36$\\
    \texttt{making scrambled \text{Eggs}} & 7.9 & $4.0e+30$ \\
    \texttt{making a \text {bonfire}} & 8.0 & ${3.5e+20}$ \\
    \texttt{making a \text{coffee}} & 8.0 & $9.8e+21$\\
    % \texttt{ordering a \text{pizza} } & 5.7 & $2.4e+31$\\
    \texttt{paying with a credit card} & 7.8 & $2.4e+21$ \\
    \texttt{planting a \text{Tree}} & 3.7 & $1.6e+{16}$ \\
    \texttt{playing \text{Tennis} } & 6.7 & $1.1e+18$\\
    \texttt{renovating a \text{room}} & 8.3 & $3.1e+31$ \\
    \texttt{repairing flat \text{bicycle} Tire} & 3.4 & $8.4e+{18}$ \\
    \texttt{riding on a \text{bus}} & 3.8 & $1.0e+{17}$ \\
    \texttt{sewing a \text{button}} & 7.5 & $7.7e+28$\\
    \texttt{taking a \text{bath}} & 3.7 & $3.1e+{27}$ \\
    \texttt{taking a \text{shower}} & 7.6 & $2.2e+30$\\
    \texttt{taking a driving \text{lesson}} & 7.9 & $3.2e+15$\\
    \texttt{taking a child to \text{bed}} & 3.7 & $4.4e+15$\\
    % \texttt{taking the \text{underground} } & 9.5 & $1.4e+24$\\
    \texttt{washing ones \text{hair} } & 7.4 & $8.8e+34$\\
    \texttt{washing dishes} & 7.6 & $7.3e+27$\\
    \bottomrule
    \end{tabular}
    }
    \caption{The table provides details of the generated graphs for \scenariocount scenarios.}
    \label{tab:graph_details_number_of_paths_degree}
    \vspace{-5mm}
\end{table}

% \begin{figure}[!t]
% \begin{center}
% \includegraphics[width=\linewidth]{./images/compact_graphs/baking a cake.pdf} 
% %\includegraphics[width=\linewidth]{./images/compact_graphs/baking a cake.pdf} 
% \caption{The figure shows the generated graph for the scenario \texttt{``baking a cake''}.}
% % \label{fig:compact_graphs_changing_batteries}
% \label{fig:compact_graphs_baking_a_cake}
% \end{center}
% %%\vspace{-7mm}
% \end{figure}

\noindent\textbf{Annotations and Event Alignments:} Though there can be multiple ways (various ESDs) of describing script for a scenario, there exists an alignment among events in multiple descriptions.  The alignments assign generic groups to an event. For example, events like \texttt{``go inside the car,'' ``get into your car,'' ``enter the car,''} etc., are assigned a group like \texttt{``get-into-car.''} DeScript provides these alignments between the events for only $10$ scenarios. We extend these alignment annotations and perform the annotations for all $40$ scenarios. A group of 3 annotators (graduate students) performed all the annotations as a part of a course research project. Annotators were asked to make generic clusters to perform the specific task and assign each event to these clusters. It took around 4-12 hrs (spanning across a month) for an annotator to complete annotations and alignments for a scenario. The varied amount of time highlights the task complexity and variety in descriptions. Further, we manually inspected the alignments and found the quality of the 3 scenarios to be too noisy, and we discarded these. Hence, only 37 were used in the final analysis. 
% \AJ{have to change accordingly, we can mention that we did it for 30 scenarios and found the quality of 3 to be bad/messy, hence only 37 were used in the final analysis}.
% \AJ{some comments about the challenges of inter-annotator agreement and manual validation after alignment would be good to add here since the reader does look for inter-annotator agreement whenever a human annotation section is proposed. We can add the challenges mentioned in the rebuttal if required (have to discuss about it).}

% \AMC{\st{I think we can move the entire remark para to Appendix and in the above para mention: ``Please refer to App. X for details about automated evaluation of alignments. "}}\AJ{this was added due to multiple weaknesses pointed out by the reviewers, would be better if we keep it}

\noindent \textbf{Remark:} Unlike classification tasks, clustering doesn't typically have predefined categories. This makes it harder to establish a common framework for agreement between the annotators. Moreover, clustering comparison is challenging, and though there are metrics for comparing clusterings (e.g., Rand Index \cite{randindex}, Adjusted Mutual Information \cite{adjusted-mutual-information}, Fowlkes–Mallows index \cite{Fowlkes–Mallows-index}, etc.), a robust widely accepted metric for annotator agreement in clustering tasks is not readily available. It is to be noted that we create the dataset by sampling trajectories from the created DAG, which shows a way of performing the entire activity. Hence, we used the same to assess the quality of the annotations and made suitable changes by manual inspection. Note that although the clustering annotations may vary (in terms of granularity), the final task (defined in the later section) is only dependent on the trajectory sequence, making it suitable for the generated commonsense queries.

% \AMC{I think the process described in the next para is very similar to what was there in ScriptWorld, to save space we can describe it very succinctly and move most of the things to Appendix or refer the ScriptWorld paper.}

\noindent\textbf{Graphical Representation:} Taking inspiration from \citep{scriptworld-ijcai2023}, we create a graph structure (also referred to as Scenario Compact Graph or Compact Graph for short) from event alignments. In the graphical representation, each cluster (group) is a node in the graph. For connecting the nodes with directed edges, we use the original description sequence provided in the ESDs. In particular, a directed edge is drawn from node $p$ to $q$ if there is at least one action (telegram-style step description) in node $p$ that directly precedes an event in node $q$. This simple strategy leads to a rich graph structure of scenarios that resembles the human understanding of these tasks. Fig. \ref{fig:compact_graphs_baking_a_cake} shows an example of such a graph. These directed acyclic graphs (DAGs) provide a medium for generating enormous trajectories 
% (
% scales from $1.6e+{16}$ to $2.6e+{30}$ \AJ{Recheck the ranges once}, 
% also 
(refer to Table \ref{tab:graph_details_number_of_paths_degree}), that are coming directly from human annotations (alignment annotation as well as the ESDs written by crowd-soured workers), providing us a proxy to represent the understanding of daily activities. 
% \st{Table (tab:graph-details-number-of-paths-degree) highlights the huge number of ESDs obtained via combinations with the limited number of available ESDs.} 
% \AJ{move later to the appendix}
% This essentially captures the generative aspect of languages, where one learns a task by performing it a few times, and the same task can be described in a massive number of combinations of generic steps. Each node in the presented graph also contains miniature steps. For example, for the subtask \texttt{``take medicine''} (represented by a single node in the entire graph), some crowdsource workers explain it in more detail, like \texttt{``open water bottle,''} \texttt{``put medicine in the mouth,''} \texttt{``drink water.''} To handle such cases, we expand the node further. 
We provide more details about graphical representations and computing the total number of ESDs in the App. \ref{sec-app:number-paths}. 
% \st{Note that the total number of ESDs that can be generated using the created graph is the total number of paths/trajectories from the start node to the end task node. (see Fig. (fig:compact-graphs-baking-a-cake) for reference)}\AMC{No need to mention previous line in here, have it in appendix.}
% We compute the number of parallel paths in the compact graph first (using DFS) and then include the paths in the created scenario graph (by incorporating the parallel paths for all sub-tasks). 

\noindent \textbf{Trajectory Entropy:} To quantify the complexity across various scenarios and compare the created graphical representations in detail, we also define Trajectory Entropy $\mathcal{H}_t$ (details in App. \ref{sec-trajectory_entropy}). Fig. \ref{fig:trajectory_entropy_vs_num_of_paths} provides a comparison of various scenarios in terms of number of paths and the defined Trajectory Entropy $\mathcal{H}_t$. 
% \AMC{maybe can we move this to appendix} 
% We find that though there is a relationship between the entropy with the number of paths. \st{there are a few outliers like \texttt{`playing tennis'}, \texttt{`ironing laundry'}, and \texttt{`renovating a room'}, and the entropy would be another measure to identify the complexity of the task captured by the compact graph representations.} \AMC{No need to mention previous line in here, have it in appendix.} 

\noindent\textbf{Reasoning Question Creation:} 
To test LLMs for commonsense knowledge understanding, we would like to generate commonsense reasoning questions related to the obtained activities. We generate questions via compact graphs. Each path in the compact graph denotes a suitable set of steps (events) for accomplishing a task. Using the graph, we sample multiple trajectories for finishing the task $t_1, t_2, \ldots, t_n \in \tau$. Each of these trajectories contains multiple events  
% \AM{\st{there is confusion about steps and events, this should be made clear or just stick to events}} 
of ESDs, $e_1, e_2, \ldots e_{m_{t_{i}}}$. Note, since different trajectories may require different numbers of steps, $m_{t_{i}}$ (referred to as $m$ when it is clear from the context) is a random variable here, which depends on the selected trajectory $t_i$. Given a trajectory, we further use a subpart of the trajectory by taking a split at a step $n \in \{1, m\}$ and use steps $e_1, e_2, \ldots e_{n-1}$ as a part of a commonsense reasoning question and $e_n$ as the correct choice for the question. Using the obtained samples, we use a template prompt to generate a commonsense reasoning question. App. Fig. \ref{fig:prompt_template_autoregressive} shows a template prompt.

\begin{figure*}[t]
\begin{center}
\includegraphics[width=0.9\linewidth]{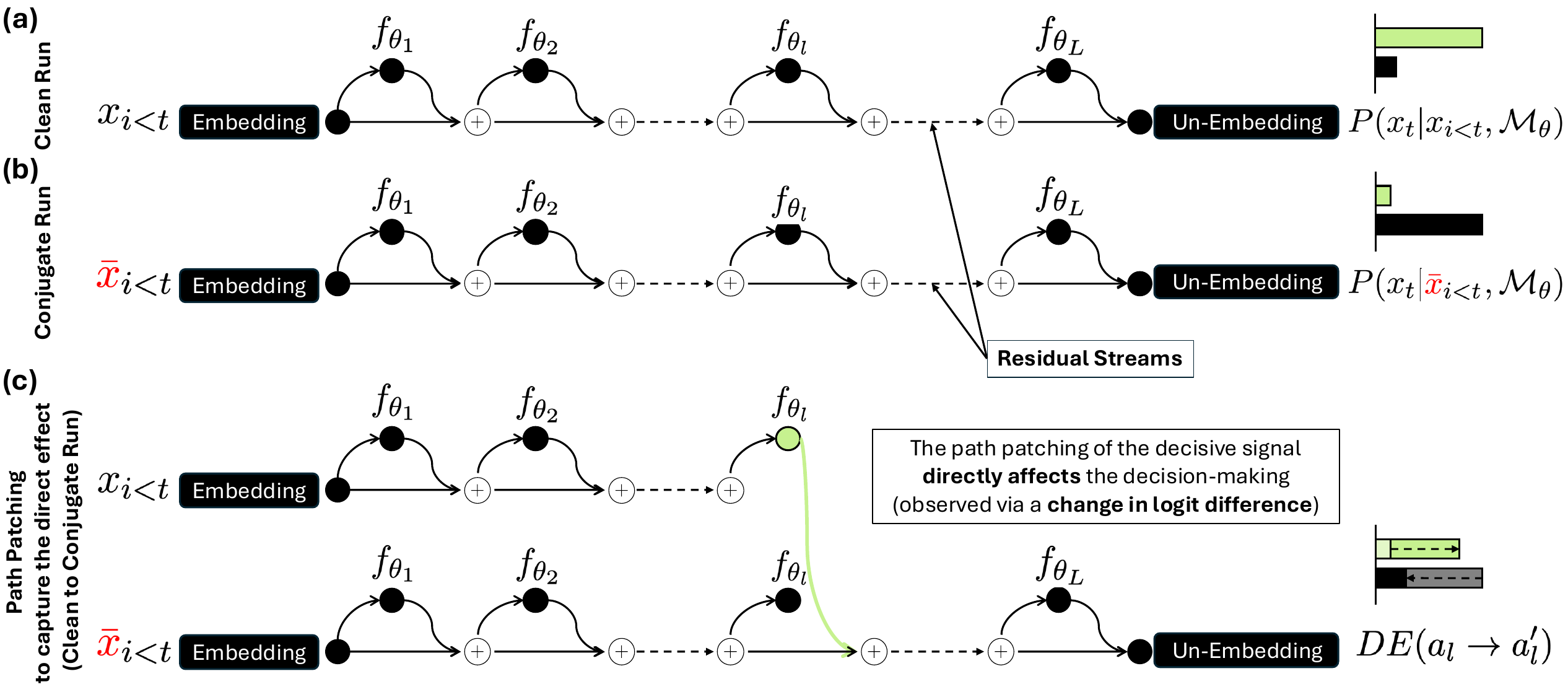} 
\caption{
The figures highlight the computation of direct effect via path patching. \textbf{(a)} A run with the clean prompt ($x_{i<t}$) is passed through the model, saving all the intermediate states. (b) A model pass is again done using a conjugate prompt (${\color{red}{{\bar{x}}}}_{i<t}$) that flips the expected behavior of the model from green option to black option. (c) A run for computing the direct effect is done, where a path patching takes place for $f_{\theta_l}$, i.e., the green signal is patched to the conjugate run. The change in logit values helps localize the decision-making component that plays a vital role in the model selecting green as the correct choice.
% \AMC{Explain the figure, also in the appendix, let's make one diagram of the actual transformer architecture with residual stream and side-by-side show figure 3a in parallel to it....this will help to understand figure 3 as it is not understandable by everyone, we can refer to this parallel figure in here for more understanding}
}
% \label{fig:compact_graphs_changing_batteries}
\label{fig:conjugate_prompts_direct_effect}
\end{center}
\vspace{-5mm}
\end{figure*}

\noindent\textbf{Data Quality Check:} 
A noteworthy point about the created dataset is that although it is generated using an algorithmic procedure, the core knowledge still comes from humans.
The algorithmic generation provides an added advantage of exhaustiveness with a meager human annotation cost, making the generated distribution of commonsense queries less likely to be previously seen by the pretrained LLMs. We additionally perform some manual inspection to improve the dataset quality (details in App \ref{app:improving_data_quality}).
% \AJ{the algorithm is not perfect, so we cross-validate}
% \AJ{To cross-validate the quality of the generated dataset, we perform additional checks of the created DAGs by manual inspection of compact graph structure (and improve the quality by manually removing the nodes/entries/edges that do not form an explainable path from the start node to the end node), manual inspection of the descriptions that are clubbed together.}
Lastly, we conducted a sanity check, where we took a sample of 1k commonsense queries for 5 of 37 scenarios and asked 5 human annotators to know how well humans perform on the created task. We recorded an average accuracy of $95\%$ with $92\%$ and $98\%$ being the minimum and maximum, respectively (more details in Table \ref{tab:human_performance}), 
% \AJ{minor comment by a reviewer}, 
validating the commonsense captured by the created queries. Interestingly, we ran an evaluation over the same set of 1k queries using one of the proprietary-LLM (\texttt{`claude-3.5-sonnet-20240620'}) and observed a success rate of 94.30\%, which is very close to human performance. 
% \AMC{We need to briefly mention (3-4 lines in here and rest in appendix as part of related work): how is our dataset diff from other commonsense datasets and what advantages does it have; e.g., exhaustive, daily scenarios, facilitates designing of experiments that help to understand reasoning mechanism in LLMs.}

% \AJ{
To this end, the proposed scheme can create an enormous number of commonsense queries ($\sim10^{17}$ for a single activity), facilitating a rigorous/exhaustive evaluation of commonsense knowledge about these activities.
% Note that though the proposed scheme is capable of generating enormous queries, 
% we perform all the analysis on the dataset generated from 2k trajectories (leading to $\sim 20k$ commonsense queries) for each scenario. We freeze this dataset of ($\sim 20k$ commonsense queries per scenario) for easier replicability of the obtained results.

% \textbf{Note:} \AJ{can add a detailed remark for the usage of the terms exhaustive and rigorous}
% \newpage
\noindent\textbf{Remark on Terminology:} We use the word "exhaustive" specifically in reference to the procedural knowledge captured in the DeScript corpus, which denotes comprehensive coverage of event orderings and dependencies for the human-authored activities in our framework, not universal commonsense knowledge which varies culturally as well as contextually. The proposed scheme enables testing over enormous trajectories per activity, exhausting the solution space defined by the original crowd workers’ procedural annotations, making it a suitable proxy for capturing the underlying commonsense knowledge in these activities.

\section{
% Mechanistically Dissecting the 
% Neural Mechanisms of Commonsense Reasoning \AJ{section name too long}
A Tool for Mechanistic Insights
}

In recent times, pretrained transformer-based networks have shown remarkable performance in a wide range of tasks \cite{devlin-etal-2019-bert, Radford2019LanguageMA, incontextfewshotlearners}, including some of the popular commonsense reasoning tasks \cite{zellers2019hellaswag,zhao2023largelanguagemodelscommonsense}. However, the understanding of decision-making happening inside these large models remains limited. 
With the help of the proposed dataset generation scheme, we would like to investigate how a commonsense reasoning query is answered by these large decoder-only autoregressive transformer-based language models autoregressive transformer models. 
% \AMC{\st{instead of ``large GPT-like autoregressive transformer models" maybe write it as ``large decoder-only autoregressive transformer-based language models"}}. 

Though there have been some works localizing the information in these models \cite{wang2022interpretabilitywildcircuitindirect,meng2023locatingeditingfactualassociations,goldowskydill2023localizingmodelbehaviorpath},
tools to tease apart the decision-making happening inside these models remain limited. 
% \AJ{In recent years, a wide range of approaches have been proposed in the context of factual recall\AM{cite}, where the recalling circuit for a particular fact is found via circuit-level attribution in language models. A representative work widely used across these methods is the counteract dataset \AM{cite}, which provides flexibility in choosing the counterfactual statement. Specifically, the dataset consists of a series of prompts and combines a tuple (subject, relation, object), and the object is replaced by a counterfactual object, making sense in the context. This helps tease apart the factual recalling mechanism by producing prompts whose completion requires specific factual knowledge about a subject and a relation.}
% In this work, 
We investigate if the decision-making in these commonsense reasoning queries can be localized. 
% \AMC{last line looks a bit isolated from the remaining paragraph, which is about factual recall. Not sure if we really need to go into details of the whole factual recall thing and counteract the dataset; this seems more suitable for related work.}

\noindent A prompt acting as an input to a Language Model (LM) comprises information related to the query that helps determine the expected answer. 
In our setup, we focus on the multiple-choice question answering (MCQA) prompt, which consists of two critical components, \textbf{1) Incomplete Task Trajectory} ($traj.$): which includes the sequence of states or steps, capturing the partial progression toward completing the task. \textbf{2) A Choice Set} 
% \AM{
(\textbf{A.}\  $o_{correct}; \ \textbf{B.}\ o_{wrong}$)) consisting of two options from which the LM must select the correct answer and generate as output either \textbf{A} or \textbf{B}. Note that the \textbf{A.} and \textbf{B.} are for representation, and in the actual run, the correct/wrong options are shuffled to marginalize the effect of models choosing a specific option.
% }. 
% \AJ{explain shuffling of options for better results}
% \AMC{we need to make it consistent with MCQA format that we are actually using. We can also explain shuffling here. This avoids confusion later in this section.}

% \st{\textbf{2) A Choice Set}  ($o_{correct}, o_{wrong}$): consisting of two options from which the LM must select the correct answer.}

% \AM{
The decision taken by the LM (($\mathcal{M}_\theta$), where $\theta$ represents the model parameters) depends on these two critical components.
% } \st{along with the language model ($\mathcal{M}_\theta$), where $\theta$ represents the model parameters.} 
Additionally, the predictions also depend on the way in which the query is framed, i.e. the prompt template ($x_{\epsilon}$) used to frame the queries.
The predicted probability/logit value of the next token can be written as
\begin{align*}
P(x_t|x_{i<t}, \mathcal{M}_{\theta}) &= P(x_t|x_{traj.} , x_{options} , x_{\epsilon}, \mathcal{M}_{\theta})\\
x_{traj.} &\leftarrow \{s_1, s_2, \ldots, s_n\}\\
x_{options} &\leftarrow \{
% \AM{
\text{\textbf{A. }}
% }
o_{correct}, 
% \AM{
\text{\textbf{B. }}
% } 
o_{wrong}\}\\
x_{\epsilon} &\in \text{set of prompt  templates} \\ 
\mathcal{M}_\theta &= \{f_{\theta_1}, f_{\theta_2}, \ldots f_{\theta_L}\}
\end{align*}
% 
%\AMC{In the above notations, is it a single template or multiple templates? as you have written ``set of prompt template\textbf{S}"}
%\AMC{i think we need to write about the A. and B. thing in here itself, so that there is no confusion later} 
% 
In the transformer-based language model, the input prompt ($x_{i<t}$) is passed through a sequence of transformer blocks/layers ($f_{\theta_1}, f_{\theta_2}, \ldots f_{\theta_L}$), providing a distribution of logits over the vocabulary for the next tokens, we only consider the predicted distribution of the last token ($x_t$), i.e., the token responsible for answering the reasoning query (using logits corresponding to tokens `\underline{ }A' and `\underline{ }B', see Fig. \ref{fig:prompt_template_autoregressive} for reference).
% ($x_{i+1<t+1}$) \AMC{the subscript should be } given previous tokens ($x_{i<t}$). \AJ{keep only the last token and explain the actual thing in detail in the appendix (experimental details)}
\begin{align*}
% x_{i+1<t+1} &= \mathcal{M}_\theta(x_{i<t}) \\
\mathcal{M}_\theta(x_{i<t}) &= f_{\theta_L}(1 + f_{\theta_{L-1}}( \ldots (1 + f_{\theta_1}(x_{i<t})))
 % \\
% \mathcal{M}_\theta(x_{i<t}) &= f_{\theta_L} \circ (1 + f_{\theta_{L-1}}) \circ  \ldots \circ (1 + f_{\theta_{1}} (x_{i<t}))\\
\end{align*}
% 
% \AMC{\st{Above, we do not need second equation}}
\noindent These sequences of operations play a crucial role in modifying the residual stream (the $1 +$ denotes the update in the residual stream throughout the transformer blocks), leading to the final predicted token $x_t$. Fig. \ref{fig:conjugate_prompts_direct_effect} (a) highlights the signal passing through the residual stream where transformer blocks are present in parallel. 
Note, in some of the transformer implementation designs, there are two points in a single transformer block where the computational blocks read/write back from/to the residual stream (self-attention and MLP); we skip the mid-skip connection in the equations above for brevity.
%  \AJ{part a of the Fig. can be referred} 
% \AMC{I think we need to briefly explain what is residual stream with a diagram? Possibly we mention it here and include details in the appendix}
% 

\noindent \textbf{Direct Effect: } To measure the effect of the transformer’s $l_{th}$ layer over the predicted decision, we make use of the direct effect, we follow \citet{Chattopadhyay2019NeuralNA, meng2023locatingeditingfactualassociations,mcgrath2023hydraeffectemergentselfrepair,heimersheim2024useinterpretactivationpatching} assuming the transformer-based architectures as structural causal models (SCMs) \cite{pearl2016causal}. 
% \AMC{need to explain very briefly what is SCM or we can point to a reference to explain what is SCM for people who are new to causality?} 
The direct effect of intervening over the activations $A_l = a_l \rightarrow A_l = {{a}^{\prime}}_l$ is computed as
\begin{align*}
& DE(a_l \rightarrow a_l') = \\  & \qquad \qquad P(x_t \mid do(A_l = a_l', A_{\neq l} = a_{\neq l}(x_{i<t}))) \\& \qquad \qquad - P(x_t \mid do(A_l = a_l(x_{i<t})))
\end{align*}
% 
% \begin{align*}
% DE(a_l \rightarrow a_l') &= P(x_t \mid do(A_l = a_l', A_{\neq l} = a_{\neq l}(x_{i<t}))) \\
% &\quad - P(x_t \mid do(A_l = a_l(x_{i<t})))
% \end{align*}

% \begin{align}
% DE(a_l \rightarrow a_l') &= P(x_t \mid do(A_l = a_l', A_{\neq l} = a_{\neq l}(x_{i<t}))) \nonumber \\
% &\quad - P(x_t \mid do(A_l = a_l(x_{i<t})))
% \end{align}

% $$
% DE(a_l \rightarrow a_{l}^{\prime}) = P(x_t|do(A_l={{a}^\prime}_l, A_{\neq l}=a_{\neq l}(x_{i<t}))

% - 
% P(x_t|do(A_l=a_{l}(x_{i<t})) 

% $$
where $do(.)$ denotes the do operator \cite{do_calculus} showing the intervention on $A_l$, i.e.,
estimating the effect of intervening at the $l_{th}$ layer’s activation $A_l$ and setting the value to ${a}^{\prime}_l$, keeping all the other activations intact $A_{\neq l}=a_{\neq l}(x_{i<t})$ to the value that they would have when passing $x_{i<t}$ as input prompt. The second term helps capture the effect, representing the model output, i.e.
$P(x_t \mid do(A_l = a_l(x_{i<t}))) = P(x_t|x_{i<t}) $.
This way of computing the intervention via replacing activations is also known as \textit{path patching} \cite{wang2022interpretabilitywildcircuitindirect,goldowskydill2023localizingmodelbehaviorpath} (also see Fig. \ref{fig:conjugate_prompts_direct_effect}).
% 
% \begin{align*}
% & DE(a_l \rightarrow a_{l}^{\prime}) = &\\ & \qquad \qquad P(x_t|do(A_l={{a}^\prime}_l, A_{\neq l}=a_{\neq l}(x_{i<t}))
% \\& \qquad \qquad - 
% P(x_t|x_{i<t}) 
% \end{align*}
% 
% \AMC{Above two equations can be merged into one} \AJ{make the last term inline}
% 
% 
Essentially, the direct effect measures how much changing the activation would affect the output logits if all other units were kept constant, i.e., in the setup of a language model, only units that are connected via the residual path to the output can have a direct effect. 

% \AMC{can we also show a diagram in this regard?}

% \AMC{we need to introduce the term ``path patching", so that it becomes easier to understand later, e.g., in Fig. \ref{fig:conjugate_prompts_direct_effect}}

\noindent \textbf{Intervention with Corrupted run:} 
A crucial aspect of capturing the direct effect is the choice of clean and corrupted runs. A clean run denotes the expected behavior. In contrast, a corrupted run signifies changes in the inputs that disrupt/deviate the expected behavior. To localize the decision-making happening in the network parameters, we take a corrupted run and intervene over the activations via 
% \AMC{is it ``intervene in" OR ``intervene via"} 
representations coming from the clean run. We further observe which interventions restore the expected behavior, highlighting the components that play a vital role in commonsense reasoning. Another common, widely used strategy is to patch the clean run over the corrupted run, where a Gaussian Noise is added to the same clean input (also known as Causal Tracing \cite{meng2023locatingeditingfactualassociations}). Some of the previous works \cite{heimersheim2024useinterpretactivationpatching} highlight the significance of constructing a corrupted run via similar prompts (or counterfactual prompts), making them more decisive in comparison to other methods. The flexibility in the proposed framework of commonsense queries coming from a DAG opens up a wide scope for constructing such queries.
% \AMC{\st{there might be redundancy in the above para, e.g., not sure, do we need to describe Causal Tracing in here.}}\AJ{the causal tracing is a widely used method for localization, and we mark it as one of our baselines for the patching using conjugate prompts that we propose in this paper}

\noindent \textbf{Conjugate Prompts:}
To be more decisive in the decision-making via path patching. We define a new way of constructing the corrupted run prompts. We call these Conjugate Prompts. 
For any query prompt ($x_{i<t}$), we can construct a conjugate query prompt by replacing the trajectory tokens with trajectory where the wrong option becomes the correct choice and vice versa, keeping the set of choices in the prompt intact. App. Fig. \ref{fig:conjugate_prompt_template_autoregressive} provides a pair of conjugate prompt templates. 
% \AJ{ref Fig. 11 } \AMC{include an example possibly in the appendix for this} 
This strategy helps capture the specific dependency on the trajectory, and after sampling multiple such trajectories, one could be more decisive about the localization of decision-making in the clean trajectory. 
Note that the constructed query consists of multiple segments
\begin{align*}
P(x_t|x_{i<t}, \mathcal{M}_{\theta}) &= P(x_t|x_{traj.} , x_{options} , x_{\epsilon}, \mathcal{M}_{\theta})\\
x_{traj.} &\leftarrow \{s_1, s_2, \ldots, s_n\}\\
x_{options} &\leftarrow \{ {\text{\textbf{{A. }}}} o_{correct}, {\text{\textbf{{B. }}}} o_{wrong}\}\\
x_{\epsilon} &\in \text{set of prompt  templates}
\end{align*}
and the $o_{correct}$ is the $s_{n+1}$ whereas the $o_{wrong}$ comes from a randomly sampled node (far from the current node) of the compact graph 
% \AMC{we need to take care of A. and B. thing in here}
. For the construction of a corrupted prompt that provides a decisive distinction, one would need a prompt that flips the answer. We create such prompts by taking the $o_{wrong}$ and sample a conjugate trajectory that starts at the start node and ends at the wrong node ($o_{conjugate} \leftarrow o_{wrong}$). 
We further construct the conjugate prompt (${\color{red}{\bar{x}}}_{i<t}$) by replacing the $x_{{traj.}}$ with $x_{\color{red}{\bar{traj.}}}$. 
\begin{align*}
% x_{i<t} &= x_{{traj.}} +  x_{options} + x_{\epsilon}\\
{\color{red}{\bar{x}}}_{i<t} &= x_{\color{red}{\bar{traj.}}} +  x_{options} + x_{\epsilon}\\
% x_{traj.} &\leftarrow \{s_1, s_2, \ldots, s_n\}\\
% x_{options} &\leftarrow \{o_{correct}, o_{wrong}\}\\
% x_{\epsilon} &\in \text{set of prompt  templates}\\
% \mathcal{M}_\theta &= \{f_{\theta_1}, f_{\theta_2}, \ldots f_{\theta_L}\}
P(x_t|{\color{red}{\bar{x}}}_{i<t}, \mathcal{M}_{\theta}) &= P(x_t|x_{\color{red}{\bar{traj.}}} , x_{options} , x_{\epsilon}, \mathcal{M}_{\theta})
\end{align*}
Note that the original clean run still remains the same with the same set of options present in the prompt.
\begin{align*}
x_{i<t} &= x_{{traj.}} +  x_{options} + x_{\epsilon}
% \\
% {\color{red}{\bar{x}}}_{i<t} &= x_{\color{red}{\bar{traj.}}} +  x_{options} + x_{\epsilon}\\
% x_{traj.} &\leftarrow \{s_1, s_2, \ldots, s_n\}\\
% x_{options} &\leftarrow \{o_{correct}, o_{wrong}\}\\
% x_{\epsilon} &\in \text{set of prompt  templates}\\
% \mathcal{M}_\theta &= \{f_{\theta_1}, f_{\theta_2}, \ldots f_{\theta_L}\}
\end{align*}
This minimal control helps flip the decision of a language model as for the conjugate prompt, the conjugate (wrong for clean) becomes the right choice. The direct effect of path patching on the $l_{th}$ layer, from clean run to conjugate run will be 
\begin{align*}
& DE(a_l \rightarrow a_{l}^{\prime}) = \\ & \qquad \qquad P(x_t|do(A_l={{a}^\prime}_l, A_{\neq l}=a_{\neq l}({\color{red}{\bar{x}}}_{i<t}))
\\& \qquad \qquad - 
P(x_t|{\color{red}{\bar{x}}}_{i<t}) 
\end{align*}
where $a_{l}^{\prime}$ comes from the clean run, and the remaining activations are set from the conjugate run ($A_{\neq l}=a_{\neq l}({\color{red}{\bar{x}}}_{i<t})$).
Fig. \ref{fig:conjugate_prompts_direct_effect} highlights the overall mechanism in detail, where the clean run predicts the green option being correct, whereas, for the corrupted, the model predicts the option highlighted using a black bar. Further, intervening in the signals via path patching from the clean run to the corrupted run shows the expected clean behavior (green being higher) when a decisive signal is patched from the clean run.
To capture the decision-making process, we monitor the deviations in the logits of the predicted options (i.e., the logits corresponding to `\underline{ }A' and `\underline{ }B' tokens).
% \AMC{here A. and B. will make more sense as we have already introduced it earlier}
Given the flexibility of sampling multiple such prompts, a more conclusive result about the localization of decision-making can be made.
\section{Experimental Setup: Evaluating LLMs}
%\vspace{-2mm}

We experiment with multiple (6) open-weight autoregressive models that are widely used by the community. We specifically make use of open-weight models to consider for easier replication of results and empirical transparency. Note that the primary aim of these experiments is not to benchmark the state-of-the-art models but to demonstrate the utility of the created resource for rigorous evaluation, 
% \AM{
and to enable interpretability studies in regard to commonsense understanding in LMs.
% } 

\begin{figure}
    \centering
    \includegraphics[width=0.95\linewidth]{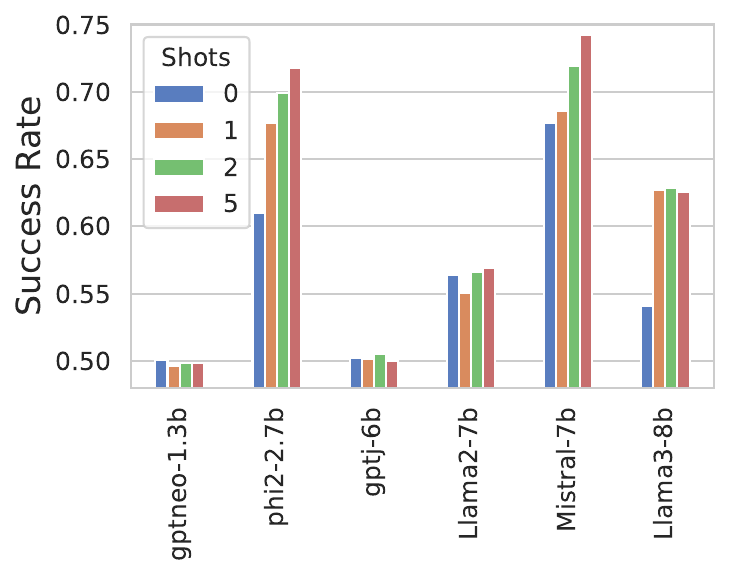}
    \caption{Success rates of different models compared across the number of shots of in-context examples. 
    % \AJ{we can also try changing the plot design for better readability.}
    % We can see that each model shows a rise in performance of selecting the next action when given more number of in-context examples. Additionally, \texttt{phi2}, in zero-shot and one-shot settings, is able to outperform larger models, including \texttt{Llama2-7b} and \texttt{Llama3-8b}. \texttt{Llama3-8b} shows a surprising decrease in success rate as we increase the number of in-context examples.
    }% \AJ{color scheme for this plot can be improved, it does not go with the rest of the plots}}
    \label{fig:success-rate-vs-model-and-shots}
    % \vspace{-5mm}
\end{figure}

\noindent \textbf{MCQA based Evaluation of Open-Weight Models:}
 For a prompt-based evaluation scheme, we frame the prompt as a multi-choice question answering (MCQA) objective \cite{robinson2023leveraging}. The prompt is intentionally structured so that the LLM is intended to predict a single-choice token (Such as `A,' `B,' etc.). \citet{robinson2023leveraging} highlight the advantages of MCQA-based evaluation over cloze evaluation \cite{incontextfewshotlearners} (where the LLMs are expected to generate the entire answer in a cloze test), leading to a significant boost in performance over various tasks, including commonsense-based tasks. Fig. \ref{fig:prompt_template_autoregressive} shows prompt templates with a qualitative example of the framed commonsense reasoning query. 
 % MCQA evaluation is performed in a zero-shot setting. 
 Additionally, to validate the effectiveness of these open-weight models over the created resource, we also include additional experiments: \textbf{1) In-Context Learning}, \textbf{2) Fine-tuning} over the generated dataset, and \textbf{3)} Investigate the \textbf{generalization} between similar scenarios in detail. We provide details of these extended experimental setups in Appendix \ref{app:experimental_setup_additional}. 
\section{Results and Empirical Findings}

In this section, we provide an in-depth insight into the model’s behavior over different aspects of the created commonsense queries. 
% We start with the general performance trend across various scenarios and move toward some empirical findings in relation to the properties of compact graphs. 
% \AMC{In this section mention about the human experiment (with reference to Table in appendix) tell abt avg performance along with performance of claude; also highlight that aim is not to benchmark LLMs but rather understanding the inner workings of LLMs}

\noindent \textbf{Overall Performance:}
Table \ref{tab:zero-shot-results} shows success rates (i.e., total percentage of commonsense queries, where the LLM generates the expected correct option) 
for different models on a zero-shot task over all \scenariocount scenarios. \texttt{Mistral-7b} shows the best performance, outperforming the other models comprehensively in the majority of the scenarios. Surprisingly, we observe that \texttt{phi-2}, which is a low-parameter model, slightly outperforms it in some scenarios.
% (\texttt{Activities})
% , only closest competitor being \texttt{phi-2} which slightly outperforms it in some scenarios. 
The results are contrary to what is expected since the performance does not scale up with the number of parameters of the model, i.e., \texttt{phi-2} outperforms \texttt{gpt-j-6B, Llama-2, and Llama-3}, and despite having fewer parameters. A similar thing can be observed in the case of \texttt{Mistral-7B} better performing than \texttt{Llama-3-8B}. 
% 
% \noindent \textbf{Additional Findings:} 
Fig. \ref{fig:success-rate-vs-model-and-shots} highlights the success rates of each model across all the scenarios when prompted with zero-shot or few-shot examples of selecting the next steps in a task. We observe that \texttt{phi2-2.7b} and \texttt{Mistral-7b} show the best performance, and their performance rises as we increase the number of in-context examples. 
% Additionally, \texttt{phi2-2.7b}, despite being smaller than Llama3 and Llama2, shows better performance than the latter two with or without in-context examples. 
% 
% \noindent 
Additionally, we perform a detailed analysis of the obtained results to better understand the behavior of these models on the created commonsense queries across 37 scenarios. Due to space limitations, we discuss the remaining analysis in the App. \ref{app:additional_results}. 

\begin{figure}
    \centering
    \includegraphics[width=\linewidth]{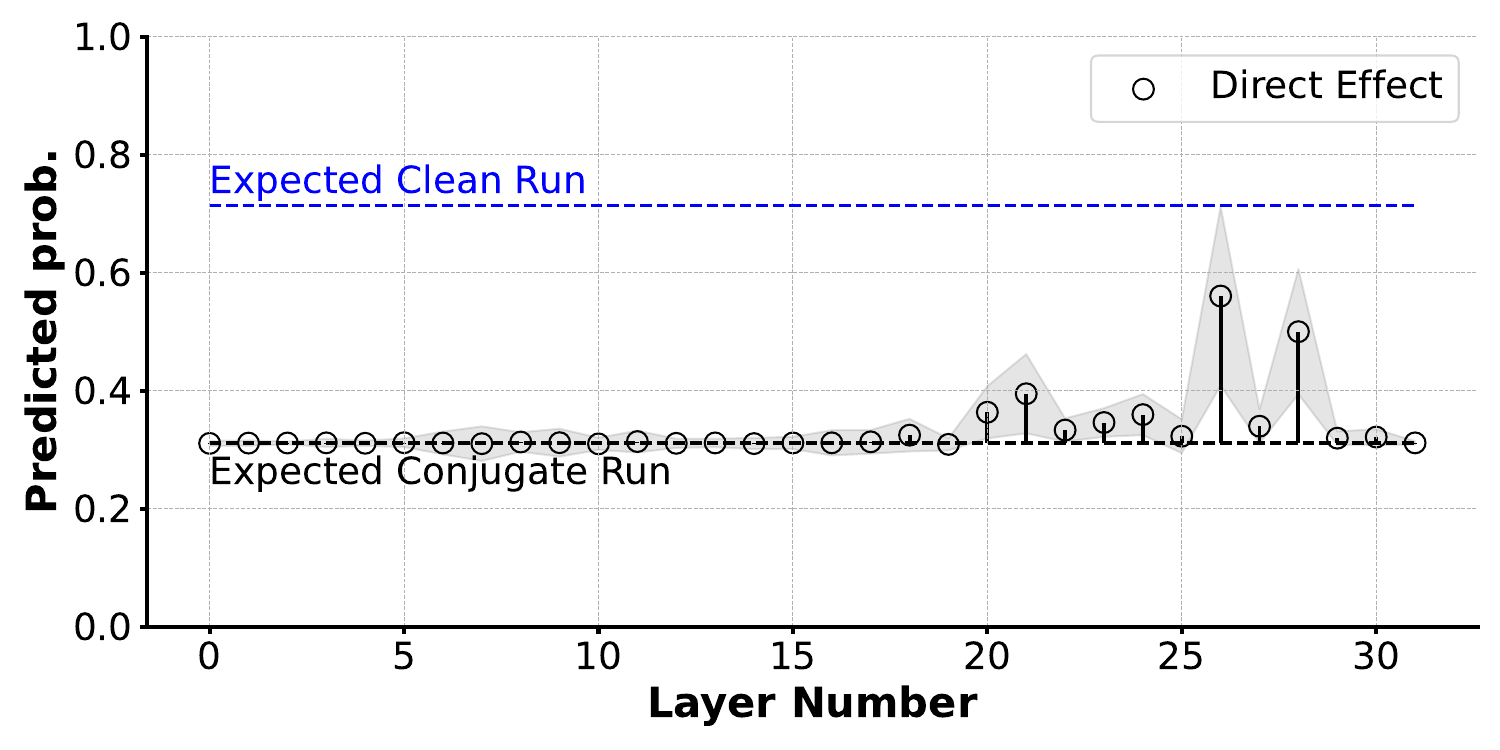}
    \caption{ The figure shows the direct effect of path patching from the clean run to the conjugate run (`\textit{going bowling}'), leading to deviations starting at layer 20 and increased signal strength at layer 26, highlighting the role of particular layers in commonsense reasoning.}
    \label{fig:decision-localization}
    % \vspace{-5mm}
\end{figure}
\begin{figure}
    \centering
    \includegraphics[width=\linewidth]{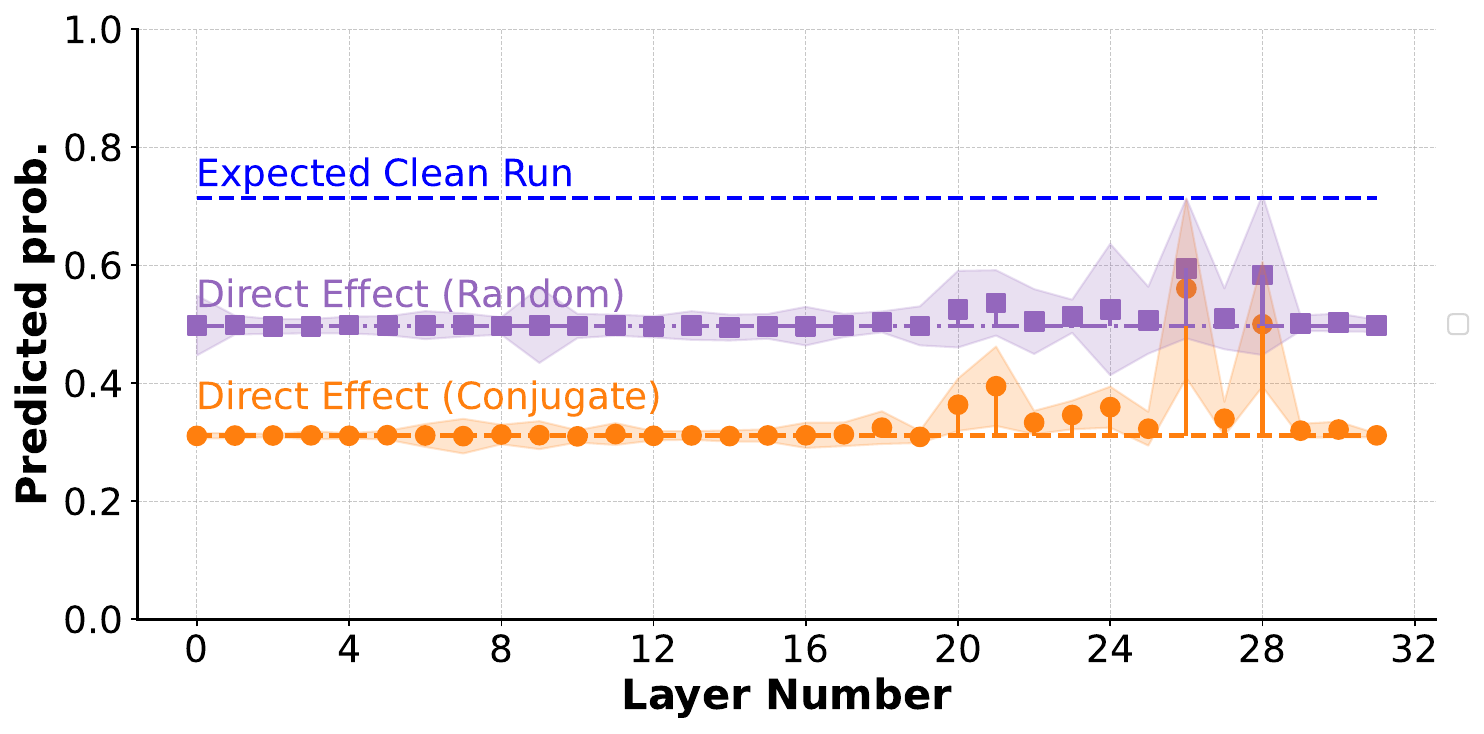}
    \caption{ The figure shows the comparison of the direct effect of path patching from the clean to the random run and clean to conjugate run (`\textit{going bowling}'). The peaks/deviations for the clean$\rightarrow$random run are less decisive than the clean$\rightarrow$conjugate run patching.
    % demonstrating the effectiveness of using the proposed conjugate prompts.
    % leading to deviations starting at layer 20 and increased signal strength at layer 26, highlighting the role of particular layers in commonsense reasoning.
    }
    \label{fig:decision-localization-comparision}
    \vspace{-5mm}
\end{figure}

Overall, we find that pre-trained \texttt{phi2} (not fine-tuned on the specific tasks) with 2.7b parameters to be providing a decent performance performance with an average of 60.67\% when compared to other models with a lower number of parameters. We choose \texttt{phi2-2.7b} to perform the localization in the decision-making experiments.

\section{Localizing Commonsense Reasoning}
% 
% \begin{figure}
%     \centering
%     \includegraphics[width=\linewidth]{./images/decision_localization_plot_random.pdf}
%     \caption{ The figure shows the direct effect of path patching from the clean run to the random run (`\textit{going bowling}'). The peaks/deviations are less decisive than \ref{fig:decision-localization}, highlighting the effectiveness of using the proposed conjugate prompts.
%     % leading to deviations starting at layer 20 and increased signal strength at layer 26, highlighting the role of particular layers in commonsense reasoning.
%     }
%     \label{fig:decision-localization-random}
%     \vspace{-5mm}
% \end{figure}

% \begin{figure*}
%     \centering
%     \includegraphics[width=\linewidth]{./images/decision_localization_plot.pdf}
%      \caption{ The figure shows the direct effect of path patching from the clean run to the conjugate run, leading to peaks at layer \AJ{add actual data}, highlighting the role of particular layers in commonsense reasoning.}
%     \label{fig:decision-localization}
% \end{figure*}
% 
To localize the components that play a primary role in the decision-making inside these models, we use the conjugate prompts (as previously explained). 
% , keeping the causal tracing experiments as our baselines
For these experiments, we consider a subset of the dataset (200 queries) for which we construct the conjugate prompts. Considering the actual performance of the phi model to be around 60\%, we only select commonsense queries where the model predicts the correct choice. 
% We initially found that there was an on-average performance shift between the conjugate prompts and the clean prompts. However, for more decisive localization experiments, balance them out by shuffling them randomly (note that the conjugate prompt relation is symmetric, and choosing any one of the two prompts (clean or conjugate)  makes no difference). We make such random shuffling to mitigate the difference between the expected logit values throughout multiple samples.
Fig. \ref{fig:decision-localization} shows the direct effect of path patching from the clean run to the conjugate run (for the 
scenario (`\textit{going bowling}')) for different transformer layers. For the initial 20 layers (layer 0 to layer 19), we observe a minimal deviation in the predicted choice from the expected conjugate run. In contrast, after 20 layers, we start observing the shift of the predicted probabilities toward the Expected Clean Run, pointing toward the patched signal being responsible for decision-making. We hypothesize layer 20 to be the primary initiator of the decision-making, and the following layers increase the strength of (or help reinforce) the decision (layer 26 to show the maximum deviation).
We perform a detailed set of these experiments over all the 37 scenarios present in the proposed framework.
Interestingly, we find that these deviations are consistent across different scenarios (see App. Fig. \ref{fig:decision-localization-all-scenarios}), and there seems to exist a few specific modules that show a peak in the direct effect, pointing towards the localization of the decision-making component present in these large autoregressive models. 
% \AJ{summary of findings from the overall results will go here}

We also observe that there is an increase in peak detection when computing the direct effect from conjugate prompts (Fig. \ref{fig:decision-localization}) when compared to a corrupted run created using a prompt with random tokens (Fig. \ref{fig:decision-localization-random}). (also see Fig. \ref{fig:decision-localization-comparision} for comparison).
% where Gaussian noise is added to the clean input for constructing the corrupted input. 
This highlights the effectiveness of the proposed conjugate prompts, making the direct effect peaks more decisive for localizing the decision-making.

\vspace{-1mm}
\section{Related Works}
\vspace{-1mm}

\noindent The proposed scheme primarily targets a special case of commonsense reasoning. In the past, a large body of research works have investigated commonsense knowledge. 
% using various methods.
Our work intersects with three broad research areas: 1) Commonsense Knowledge Resources, 2) Script-based Procedural Reasoning, and 3) Mechanistic Interpretability. 

\noindent\textbf{Commonsense Knowledge Resources:}
Some of the recent works to model commonsense reasoning include knowledge graphs like ATOMIC \cite{hwang2021cometatomic}, which captures social and physical inferences,
% through 877k tuples,
and transformer-based generators like COMET \cite{bosselut-etal-2019-comet} and the follow up works \cite{west-etal-2023-novacomet,park2020visualcomet, curiouscase,rashkin-etal-2018-event2mind}.
While these resources enable broad reasoning, they lack a granular procedural structure.
On the other hand, benchmarks such as SWAG \cite{zellers-etal-2018-swag}, HellaSwag \cite{zellers-etal-2019-hellaswag}, and COIN \cite{ostermann-etal-2019-commonsense} evaluate isolated inferences but ignore to test multi-step reasoning in procedural text. Some other works include \cite{qin-etal-2019-counterfactual, huang-etal-2019-cosmos,Bhagavatula2020Abductive,qin2021timedial,talmor2021commonsenseqa, zellers2021merlot,zhao2024uncommonsense}
Recently proposed methods show a good performance on these tasks \cite{lourie2021unicorn, zhou2023commonsense}, yet their performance remains limited to a small evaluation set, making quantification challenging. 
It is often difficult to quantify if the performance reflects surface pattern matching or structured understanding \cite{wang2024llms}. Unlike these works, our scheme models activities as directed graphs, enabling evaluation through sampling enormous trajectories per activity.

\noindent\textbf{Script-Based Commonsense Reasoning:}
Scripts have been an active area of research for the last
four decades.
Scripts provide a framework to formalize procedural knowledge as event sequences \cite{SCHANK1975237,schank1975scripts}, with corpora like InScript \cite{modi-etal-2016-inscript}, DeScript \cite{wanzare-etal-2016-crowdsourced}, and McScript \cite{ostermann-etal-2018-mcscript}, capturing commonsense knowledge via crowdsourcing. 
% Some of the early works learned script structure via neural methods \cite{modi-titov-2014-inducing} and dependency parsing \cite{frermann2014hierarchical}, while recent methods include LLMs generating scripts \cite{honovich2022instruction} or reason using counterfactuals \cite{qin-etal-2019-counterfactual}.
% As evident from the definition (\S\ref{sec:intro}), scripts encapsulate commonsense and procedural knowledge about the world and are an ideal source for training algorithms to learn about the world. 
Several computational models have developed to model script knowledge, \textit{interalia}, \cite{regneri2010learning,frermann2014hierarchical,modi-2016-event,modi-titov-2014-inducing,rudinger-etal-2015-learning,jans-etal-2012-skip,pichotta-mooney-2016-using,modi-etal-2017-modeling,lm-scripts-2021,tandon-etal-2019-wiqa,madaan-etal-2021-give,sakaguchi2021proscript, saha-etal-2021-explagraphs,li2023break,creswell2023selectioninference, Gandhi2023StrategicRW, onoe2023lms, Poesia2023CertifiedRW, cold}.
However, evaluations remain limited to small test sets and are often limited in capturing real-world variation. In this work, we expand this paradigm by converting scripts into directed graphs that encode valid event orderings per activity, supporting systematic stress-testing through marginalization over enormous trajectories.

\noindent\textbf{Mechanistically Interpretable Localization:} In recent years, a wide range of approaches have been proposed in the context of factual recall \cite{meng2023locatingeditingfactualassociations, heimersheim2024useinterpretactivationpatching, wang2022interpretabilitywildcircuitindirect, gordon-etal-2012-semeval_copa_dataset}, where the recalling circuit for a particular fact is found via circuit-level attribution in language models. A representative work widely used across these methods is the counteract dataset \cite{meng2023locatingeditingfactualassociations}, which provides flexibility in choosing the counterfactual statement. Specifically, the dataset consists of a series of prompts and combines a tuple (subject, relation, object), and the object is replaced by a counterfactual object, making sense in the context. This helps tease apart the factual recalling mechanism by producing prompts whose completion requires specific factual knowledge about a subject and a relation. 
However, most of the prior art focuses on attribute recall rather than procedural reasoning. In this work, we extend it for commonsense reasoning happening inside these large autoregressive models by providing resources that facilitate marginalization using multiple samples.

%\vspace{-2mm}
\section{Conclusion}

 In this work, we study to quantify commonsense knowledge acquired in LLMs by performing a rigorous evaluation over real-world activities well understood by humans. We provide alignment resources for \scenariocount daily human activities, which can generate an enormous number of choice-based questions for validating the commonsense reasoning in LLMs.
 % and language understanding aspect of 
 % generative models. 
 With a detailed analysis of 6 open-weight models, we find commonsense reasoning challenging for LLMs. 
 % Our generalization findings over multiple similar activities point towards LLMs learning latent representations that help facilitate knowledge transfer. 
 To add an extra layer of understanding of the performance, we dive deeper into the relationships with different properties of the scenarios and report the findings. 
 % To the best of our knowledge, this is one of its kind detailed study done for quantifying the commonsense knowledge of LLMs. 
 % We hope the generic scheme we propose will help exhaustive evaluation of LLMs in the coming future.
 Further, we provide ways in which the decision-making about commonsense reasoning happening inside these models could be localized and understood. Our analysis using the Phi-2 model points out a few localized layers that play a crucial role in predicting the expected reasonable answer. We hope that this work opens up new ways of understanding the commonsense reasoning happening inside these models, by not only grasping the representations learned by these models but also by making a comparison with the compact graph representation of the commonsense knowledge about these daily real-world activities.

 % of these latent representations, we localize the commonsense reasoning abilities in LLMs using causal mediation analysis. Our empirical findings suggest that knowledge about these activities is localized and can be transferred from one activity to another, improving performance over a new set of activities. \AJ{more will go here after the localization experiments}

\clearpage
\newpage

\section*{Limitations}

The major limitation of this work is the low number of stereotypical human activities (\scenariocount in number) used to validate the commonsense understanding aspect of LLMs. Though the validation space generated by the graph representation is enormous, the provided resource can only validate the commonsense understanding aspect in models for a limited set of these \scenariocount scenarios, which may not be the true representative of the generalized understanding activities in the wild.

Though the framework supports the flexibility of choosing a set of question prompt templates, for our experiments, given the computation cost, we find a single prompt template that shows a nominal performance and use the same for all the analyses. In the future, it would be good to marginalize the results by using multiple prompt templates.

For finding the decision-making components in the large autoregressive language models, though we provide a rich resource that facilitates teasing apart various modules. In our experiments, we only considered a small set of indicative experiments to show the utility of the proposed framework.  
Moreover, we only considered the activation blocks with less granularity, and a better localization may exist when performing path patching by analyzing the role of individual attention heads. Furthermore, we only used \texttt{phi-2} for the localization experiments, and more analysis would be required for other open-weight models that show a decent performance over the created commonsense queries.
At last, we would like to mention that these experiments only provide a weak signal that localization may exist, and the current method of direct computation may not be transparent enough to find the decision-making modules for common sense reasoning. We encourage future works to consider finding the underlying circuits behind the commonsense reasoning.
We believe the proposed framework will lead to a helpful resource with high utility, both for robust evaluation and circuit discovery of commonsense reasoning, helping find out ways in which these models can be made more accurate for commonsense reasoning in general.

\section*{Ethical Aspects}

Our work does not have any negative impact on the society. We create a dataset for evaluating LLMs for commonsense knowledge and evaluate open-weight LLMs exhaustively and rigorously.

% Bibliography entries for the entire Anthology, followed by custom entries
%\bibliography{anthology,custom}
% Custom bibliography entries only

\bibliography{references}

\clearpage
\newpage

\appendix

\section*{Appendix} \label{sec:appendix}

\appendix

%%%%%%%%%%%%%%%%%%%%%%%%%%%

\titlecontents{section}[18pt]{\vspace{0.05em}}{\contentslabel{1.5em}}{}
{\titlerule*[0.5pc]{.}\contentspage} % Set the formatting for appendix sections in the table of contents

\titlecontents{table}[0pt]{\vspace{0.05em}}{\contentslabel{1em}}{}
{\titlerule*[0.5pc]{.}\contentspage} % Set the formatting for appendix tables in the list of tables

\startcontents[appendix] % Start the table of contents for the appendix
\section*{Table of Contents} % Title for the appendix table of contents
%\addcontentsline{toc}{section}{Table of Contents} % Add the appendix table of contents to the main table of contents
\printcontents[appendix]{section}{0}{\setcounter{tocdepth}{4}} % Print the table of contents for the appendix

% \startlist[appendix]{lot} % Start the list of tables for the appendix
% \section*{List of Tables} % Title for the appendix list of tables
% %\addcontentsline{lot}{section}{List of Tables} % Add the appendix list of tables to the main list of tables
% \printlist[appendix]{lot}{}{\setcounter{tocdepth}{1}} % Print the list of tables for the appendix

\startlist[appendix]{lof} % Start the list of tables for the appendix
\section*{List of Figures} % Title for the appendix list of tables
%\addcontentsline{lot}{section}{List of Tables} % Add the appendix list of tables to the main list of tables
\printlist[appendix]{lof}{}{\setcounter{tocdepth}{1}} % Print the list of tables for the appendix

\newpage

\section{Details to Graphical Representation} \label{sec-app:number-paths}

\noindent\textbf{Graphical Representation:} 
% Taking inspiration from \citep{scriptworld-ijcai2023}, we create a graph structure (also referred to as Scenario Compact Graph or Compact Graph for short) from event alignments. In the graphical representation, each cluster (group) is a node in the graph. For connecting the nodes with directed edges, we use the original description sequence provided in the ESDs. In particular, a directed edge is drawn from node $p$ to $q$ if there is at least one action (telegram-style step description) in node $p$ that directly precedes an event in node $q$. This simple strategy leads to a rich graph structure of scenarios that resembles the human understanding of these tasks. Figure \ref{fig:compact_graphs_baking_a_cake} shows an example of such a graph. 
The created directed acyclic graphs (DAGs) provide a medium for generating enormous trajectories (scales from $1.6e+{16}$ to $2.6e+{30}$, also see Table \ref{tab:graph_details_number_of_paths_degree}), that are coming directly from human annotations (alignment annotation as well as the ESDs written by crowd-soured workers), providing us a proxy to represent the understanding of daily activities. 
% \st{Table (tab:graph-details-number-of-paths-degree) highlights the huge number of ESDs obtained via combinations with the limited number of available ESDs.} 
% \AJ{move later to the appendix}
% This essentially captures the generative aspect of languages, where one learns a task by performing it a few times, and the same task can be described in a massive number of combinations of generic steps. 
Each node in the presented graph also contains miniature steps. For example, for the subtask \texttt{``take medicine''} (represented by a single node in the entire graph), some crowdsource workers explain it in more detail, like \texttt{``open water bottle,''} \texttt{``put medicine in the mouth,''} \texttt{``drink water.''} To handle such cases, we expand the node further, incorporating such substeps. This essentially leads to an enormous number of paths/ESDs from the start to the end node in the graph.

\begin{figure*}[t]
\begin{center}
\includegraphics[width=0.6\linewidth]{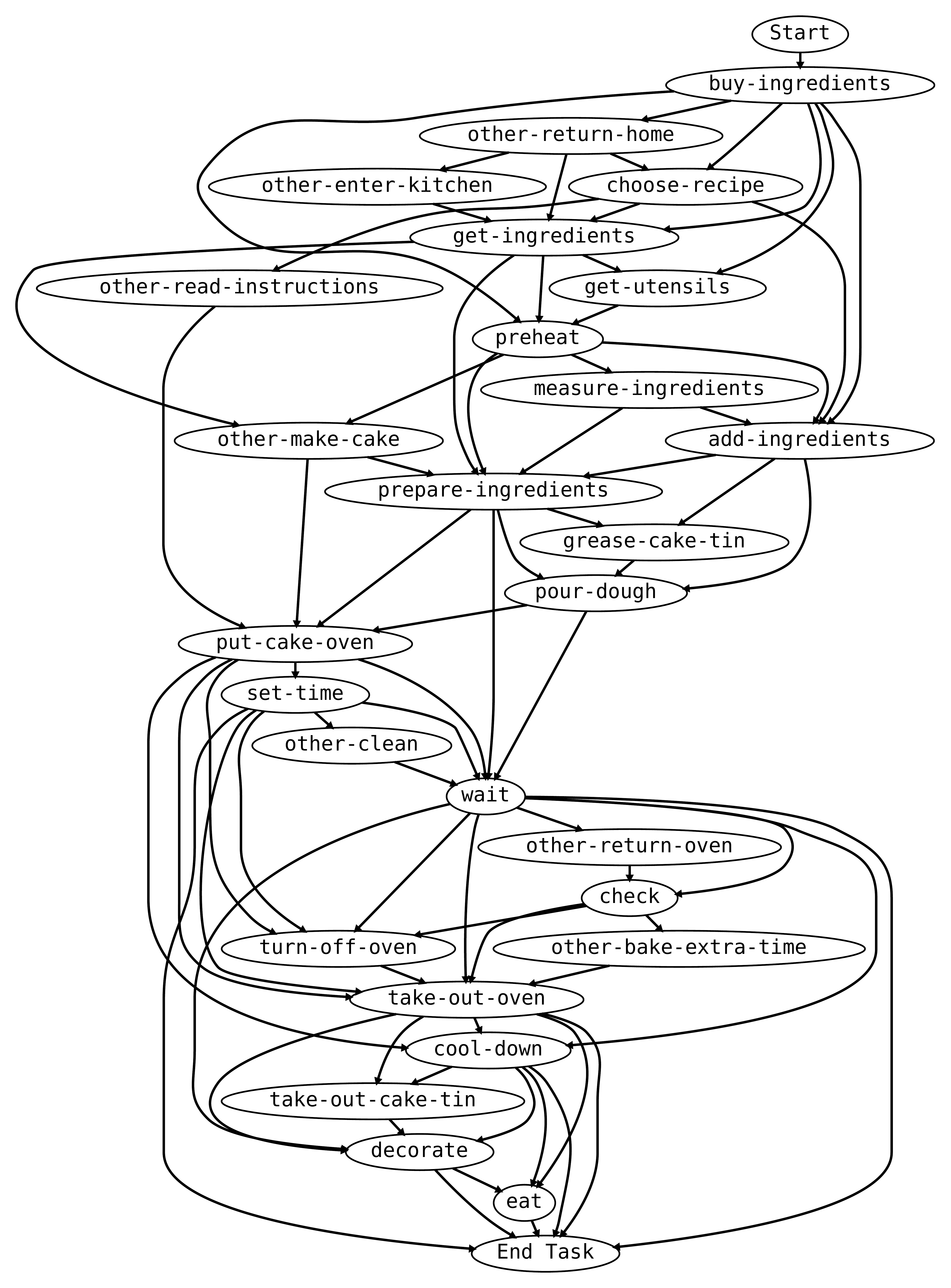} 
\caption[Compact Graph \texttt{``baking a cake''}]{The figure shows the generated graph for the scenario \texttt{``baking a cake''}.}
% \label{fig:compact_graphs_changing_batteries}
\label{fig:compact_graphs_baking_a_cake}
\end{center}
%%\vspace{-7mm}
\end{figure*}

% \AM{i think there is an overlap with the above, merge}

\noindent\textbf{Computing the total number of ESDs:} Note that the total number of ESDs that can be generated using the created graph is the total number of paths/trajectories from the start node to the end task node. (see Figure (\ref{fig:compact_graphs_baking_a_cake}) for reference)
The obtained DAGs can be used to compute the total number of paths using a simple DFS scheme. For considering the miniature steps as well, we expand the same graph 
% we compute the number of parallel paths in the compact graph first (using DFS) and then include the paths in the created scenario graph (
by incorporating the parallel paths for all sub-tasks. Further, the total number of paths is computed considering every node in the compact graph that points to multiple text instances written by different human experts.
% ). 
% Since the overall paths in the obtained graph were huge, we mathematically validated it to approximate the number of parallel paths. For an obtained graph, assuming the average number of nodes to be $20$, the total paths would be 
% $\binom{40}{20}~10^{11}(\binom{2k}{k})$
% % $40_C_20~10^11(2k_C_k)$, 
% further, since each event can be completed in multiple (parallel) ways, assuming \text{average splits for events = 4} and \text{average path length = 10}, the total path will 
% be $\binom{40}{20} \times 4^{10} \sim 10^{17}$. 
We further use these paths to get multiple commonsense reasoning question prompts. Considering the humongous number of queries, we believe the generated examples act as a closed set that captures a proxy for the commonsense understanding related to a task.

\section{Improving Data Quality} \label{app:improving_data_quality}
% \noindent\textbf{Data Quality Check:} \AJ{
A noteworthy point about the created dataset is that although it is generated using an algorithmic procedure, the core knowledge still comes from humans.
% } 
% The algorithmic generation provides an added advantage of exhaustiveness with a meager human annotation cost, making the generated distribution of commonsense queries less likely to be previously seen by the pretrained LLMs. 
% \AJ{the algorithm is not perfect, so we cross-validate}
To cross-validate the quality of the generated dataset, we perform additional checks of the created DAGs by manual inspection of compact graph structure (and improve the quality by manually removing the nodes/entries/edges that do not form an explainable path from the start node to the end node), manual inspection of the descriptions that are clubbed together.
Lastly, we conducted a sanity check, where we took a sample of 1k commonsense queries for 5 of 37 scenarios and asked 5 human annotators to know how well humans perform on the created task. We recorded an average accuracy of $95\%$ with $98\%$ being the maximum (more details in Table \ref{tab:human_performance}). 
% \AMC{We need to briefly mention (3-4 lines in here and rest in appendix as part of related work): how is our dataset diff from other commonsense datasets and what advantages does it have; e.g., exhaustive, daily scenarios, facilitates designing of experiments that help to understand reasoning mechanism in LLMs.}

\begin{table}[t]
\small
\centering
\begin{tabular}{cc}
\toprule
Human Experts   & \multicolumn{1}{l}{Task Accuracy} \\
\midrule
% Annotator-1 & 0.98                              \\
% Annotator-2 & 0.962                             \\
% Annotator-3 & 0.9651741294                      \\
% Annotator-4 & 0.9435897436                      \\
% Annotator-5 & 0.92                              \\
% Average     & 0.9541527746                     \\
Expert-1 & 98.00                              \\
Expert-2 & 96.20                             \\
Expert-3 & 96.52                      \\
Expert-4 & 94.36                      \\
Expert-5 & 92.00                              \\ \midrule
Average     & 95.42                     \\
\midrule
claude-3-5-sonnet-20240620 & 94.30 \\
\bottomrule
\end{tabular}
\caption{Performance of multiple annotators over the selected 1k samples (200 samples for 5 scenarios) over the generated commonsense queries for 5 activities. The high performance numbers indicate the presence of valid commonsense queries, well answerable by humans.}
\label{tab:human_performance}
\end{table}

\section{Prompt Templates}

Fig. \ref{fig:prompt_template_autoregressive} shows the evaluation prompt template used for MCQA-based evaluation. The prompt is intentionally structured so that the LLM is intended to predict a single-choice token (Such as `A,' `B,' etc.). 

\begin{figure*}[t]
\centering
    \scalebox{0.85}{
    \begin{tabular}{p{1.1\linewidth}}
      \toprule
      \texttt{\textcolor{blue}{[ in-context examples (if few-shot/in-context learning experiment) ]}} \\
      \texttt{Question: For the task \textcolor{orange}{activity name}, if the following steps are already completed in order \textcolor{orange}{1. $E_1$, 2. $E_2$, 3. $\ldots$ i. $E_i$}, what should be the next suitable step for completing the task? }
      % \texttt{The following are multiple choice questions \textcolor{orange}{about electrical engineering}. You should directly answer the question by choosing the correct option.} \\
      \\
      % \texttt{Which of the following events (given as options A or B) is a plausible \textcolor{orange}{question (cause/effect)} of the event \textcolor{orange}{premise}?
      % }
      % \\
      % \texttt{Question: \textcolor{blue}{In an SR latch built from NOR gates, which condition is not allowed}} \\
      % \texttt{Options:} \\
      \texttt{A. \textcolor{blue}{\textcolor{orange}{$E_{i+1}$}}} \\
      \texttt{B. \textcolor{blue}{\textcolor{orange}{wrong choice sampled from the scenario}}} \\
      % \texttt{C. \textcolor{blue}{S=1, R=0}} \\
      % \texttt{D. \textcolor{blue}{S=1, R=1}} \\
      \texttt{Answer:\textcolor{red}{ \underline{ A}}} \\
      % \bottomrule
    \end{tabular}
    }
    \scalebox{0.85}{
    \begin{tabular}{p{1.1\linewidth}}
      \toprule
      % \texttt{Consider the activity of \textcolor{orange}{activity name}.}
      % \texttt{The following are multiple choice questions \textcolor{orange}{about electrical engineering}. You should directly answer the question by choosing the correct option.} \\
      % \\
      \texttt{\textcolor{blue}{[ in-context examples (if few-shot/in-context learning experiment) ]}} \\
      \texttt{Question: For the task \textcolor{orange}{planting a tree}, if the following steps are already completed in order \textcolor{orange}{1. `Go to garden center', 2. `Obtain seedling.'}, what should be the next suitable step for completing the task? 
      %  
      % A. , \n B. 'Find a location to plant tree' \n Answer: B. 'Find a location to plant tree' \n\n?
      }
      \\
      % \texttt{Question: \textcolor{blue}{In an SR latch built from NOR gates, which condition is not allowed}} \\
      % \texttt{Options:} \\
      \texttt{A. \textcolor{blue}{\textcolor{orange}{`Water tree'}}} \\
      \texttt{B. \textcolor{blue}{\textcolor{orange}{`Find a location to plant tree'}}} \\
      % \texttt{C. \textcolor{blue}{S=1, R=0}} \\
      % \texttt{D. \textcolor{blue}{S=1, R=1}} \\
      \texttt{Answer:\textcolor{red}{ \underline{ B}}} \\
      \bottomrule
    \end{tabular}
    }
    
\caption[Input prompt formats for the MCQA-based evaluation.]{
Input prompt formats for the MCQA-based evaluation of autoregressive open-weight models (e.g., \texttt{llama(-2)}, \texttt{GPT-J}, etc.).
The \texttt{black text} is the templated input. The \texttt{\textcolor{orange}{orange text}} is the input from the current event trajectory, where the \texttt{\textcolor{orange}{activity name}} denotes the description of the activity like \texttt{baking a cake}, or \texttt{planting a tree}.
The next-token prediction probabilities of the option IDs at the \textcolor{red}{\underline{\texttt{red text}}} is used as the observed prediction distribution. 
% Note that for all the open-weight models, we do not prepend the input text with a \texttt{bos\_token}.
}
\label{fig:prompt_template_autoregressive}
% \vspace{-6mm}
\end{figure*}

\begin{figure*}[ht]
\centering
    \scalebox{0.85}{
    \begin{tabular}{p{1.1\linewidth}}
      \toprule
      % \texttt{\textcolor{blue}{[ in-context examples (if few-shot/in-context learning experiment) ]}} \\
      \texttt{Question: For the task \texttt{{\textbf{activity name}}}, if the following steps are already completed in order \textcolor{blue}{1. $E_1$, 2. $E_2$, 3. $\ldots$ p. $E_p$}, what should be the next suitable step for completing the task? }
      % \texttt{The following are multiple choice questions \textcolor{orange}{about electrical engineering}. You should directly answer the question by choosing the correct option.} \\
      \\
      % \texttt{Which of the following events (given as options A or B) is a plausible \textcolor{orange}{question (cause/effect)} of the event \textcolor{orange}{premise}?
      % }
      % \\
      % \texttt{Question: \textcolor{blue}{In an SR latch built from NOR gates, which condition is not allowed}} \\
      % \texttt{Options:} \\
      \texttt{A. \textcolor{blue}{\textcolor{blue}{$E_{p+1}$}}} \\
      \texttt{B. \textcolor{blue}{\textcolor{orange}{$E_{q+1}$}}} \\
      % \texttt{C. \textcolor{blue}{S=1, R=0}} \\
      % \texttt{D. \textcolor{blue}{S=1, R=1}} \\
      \texttt{Answer:\textcolor{red}{ \underline{ A}}} \\
      % \bottomrule
    \end{tabular}
    }
    \scalebox{0.85}{
    \begin{tabular}{p{1.1\linewidth}}
      \toprule
      % \texttt{\textcolor{blue}{[ in-context examples (if few-shot/in-context learning experiment) ]}} \\
      \texttt{Question: For the task \texttt{\textbf{{activity name}}}, if the following steps are already completed in order \textcolor{orange}{1. $E_1$, 2. $E_2$, 3. $\ldots$ q. $E_q$}, what should be the next suitable step for completing the task? }
      % \texttt{The following are multiple choice questions \textcolor{orange}{about electrical engineering}. You should directly answer the question by choosing the correct option.} \\
      \\
      % \texttt{Which of the following events (given as options A or B) is a plausible \textcolor{orange}{question (cause/effect)} of the event \textcolor{orange}{premise}?
      % }
      % \\
      % \texttt{Question: \textcolor{blue}{In an SR latch built from NOR gates, which condition is not allowed}} \\
      % \texttt{Options:} \\
      \texttt{A. \textcolor{blue}{\textcolor{blue}{$E_{p+1}$}}} \\
      \texttt{B. \textcolor{blue}{\textcolor{orange}{$E_{q+1}$}}} \\
      % \texttt{C. \textcolor{blue}{S=1, R=0}} \\
      % \texttt{D. \textcolor{blue}{S=1, R=1}} \\
      \texttt{Answer:\textcolor{red}{ \underline{ B}}} \\
      \bottomrule
    \end{tabular}
    }
    % \scalebox{0.85}{
    % \begin{tabular}{p{1.1\linewidth}}
    %   \toprule
    %   % \texttt{Consider the activity of \textcolor{orange}{activity name}.}
    %   % \texttt{The following are multiple choice questions \textcolor{orange}{about electrical engineering}. You should directly answer the question by choosing the correct option.} \\
    %   % \\
    %   % \texttt{\textcolor{blue}{[ in-context examples (if few-shot/in-context learning experiment) ]}} \\
    %   \texttt{Question: For the task \textcolor{orange}{planting a tree}, if the following steps are already completed in order \textcolor{orange}{1. `Go to garden center', 2. `Obtain seedling.'}, what should be the next suitable step for completing the task? 
    %   %  
    %   % A. , \n B. 'Find a location to plant tree' \n Answer: B. 'Find a location to plant tree' \n\n?
    %   }
    %   \\
    %   % \texttt{Question: \textcolor{blue}{In an SR latch built from NOR gates, which condition is not allowed}} \\
    %   % \texttt{Options:} \\
    %   \texttt{A. \textcolor{blue}{\textcolor{orange}{`Water tree'}}} \\
    %   \texttt{B. \textcolor{blue}{\textcolor{orange}{`Find a location to plant tree'}}} \\
    %   % \texttt{C. \textcolor{blue}{S=1, R=0}} \\
    %   % \texttt{D. \textcolor{blue}{S=1, R=1}} \\
    %   \texttt{Answer:\textcolor{red}{ \underline{ B}}} \\
    %   \bottomrule
    % \end{tabular}
    % }
    
\caption[Formation of Conjugate prompt from a Clean prompt.]{
Formation of Conjugate prompt from a Clean prompt.
The \texttt{black text} is the template input ($x_\epsilon$), 
% where the activity name of the corresponding compact graph is used
where the \texttt{\textbf{{activity name}}} denotes the description of the activity like \texttt{baking a cake}, or \texttt{planting a tree}.
The \texttt{\textcolor{blue}{blue text}} is the clean run ($x_{traj.}$) ending at step $E_p$, making $E_{p+1}$ to be the correct choice.
The conjugate run input (\texttt{\textcolor{orange}{orange text}}) is framed from a conjugate trajectory (${\bar{x}}_{traj.}$) ending at $E_q$, making $E_{q+1}$ to be the correct conjugate choice.
Note that in both prompts (clean and conjugate), the options ($x_{options}$) remain the same, i.e., $E_{p+1}$ and $E_{q+1}$ and only the clean trajectory is changed to conjugate trajectory.
The next-token prediction probabilities of the option IDs at the \textcolor{red}{\underline{\texttt{red text}}} is used as the observed prediction distribution. 
The change in the decision is monitored via the difference in logits corresponding to tokens `{\textcolor{red}{ \underline{ A}}}' and `{\textcolor{red}{ \underline{ B}}}' before and after the activation path patching.
% Note that for all the open-weight models, we do not prepend the input text with a \texttt{bos\_token}.
}
\label{fig:conjugate_prompt_template_autoregressive}
% \vspace{-6mm}
\end{figure*}

\begin{figure*}
    \centering
    \includegraphics[width=0.75\textwidth]{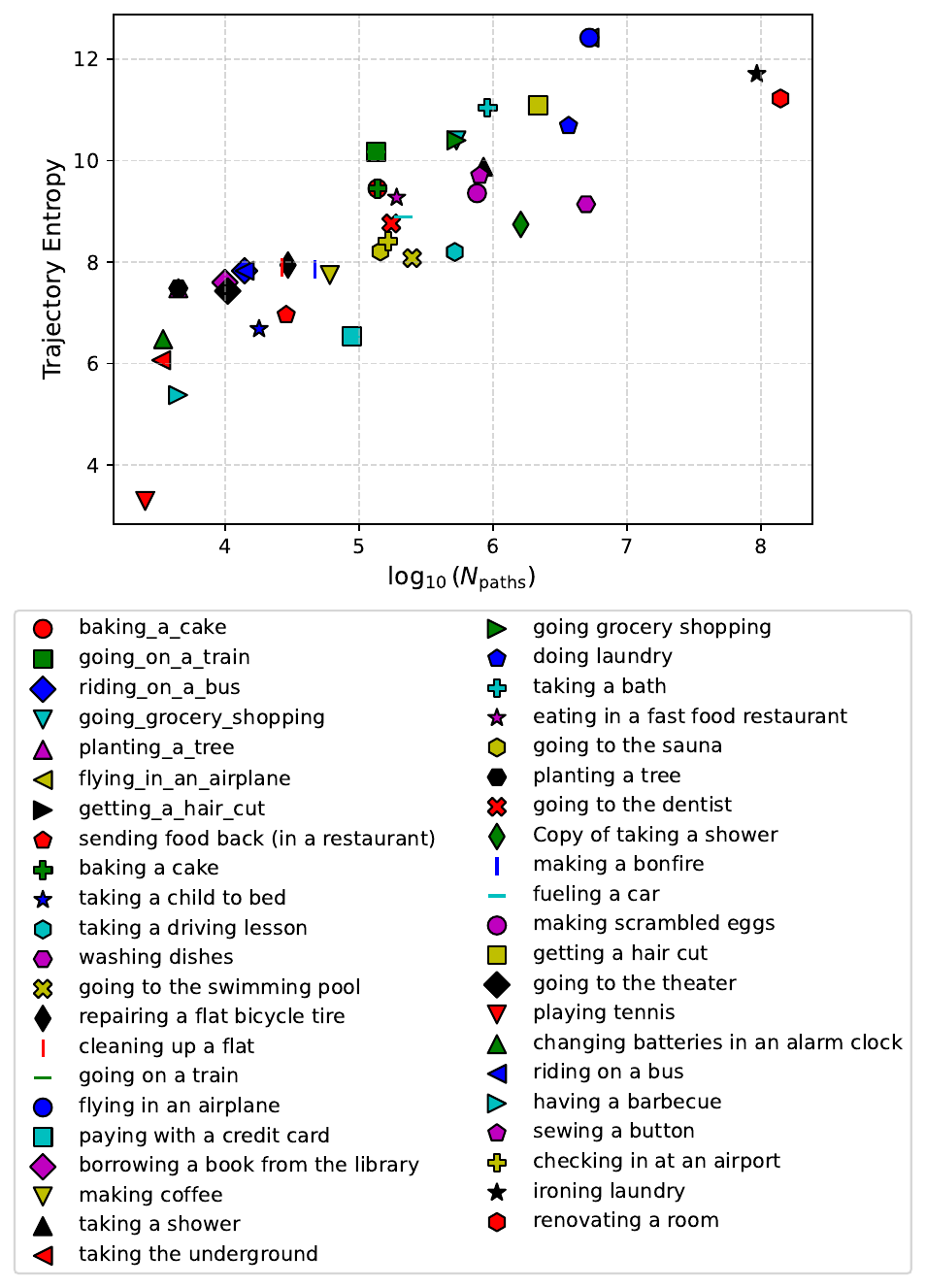}
    \caption[Variation of trajectory entropy against the number of paths]{The figure shows the variation of trajectory entropy with the Total Number of Paths in the Compact Graph. We observe that some scenarios have less or almost equal trajectory entropy despite having a significantly higher number of paths. This demonstrates that the complexity of the task is not only due to the number of paths, but some other factors also play a role in determining complexity.}
    \label{fig:trajectory_entropy_vs_num_of_paths}
\end{figure*}

\section{Trajectory Entropy} \label{sec-trajectory_entropy}
To quantify the complexity across various scenarios and compare the created DAGs in detail, we define the trajectory entropy of a scenario.
Trajectory entropy $\mathcal{H}_t$ for an DAG (Directed Acyclic Graph) 
$G$ is computed as:
\begin{equation*}
    \mathcal{H}_t = -\sum_{k=1}^N p(t_k)\text{log}\hspace{0.5mm}p(t_k)
\end{equation*}
Where $N$ is the total number of paths from the start to the end node $p(t_k)$, is the probability of trajectory $t_k$ defined as 
$p(t_k)= \prod_{ij} T (e_i\rightarrow e_j)$. $ T (e_i\rightarrow e_j)$ is transition probability from event $e_i$ to $e_j$, which is defined uniformly across all the outgoing edges. 
Figure \ref{fig:trajectory_entropy_vs_num_of_paths} shows the computations across multiple scenarios.
We find that though there is a relationship between the entropy and the number of paths, there are a few outliers like \texttt{`playing tennis'}, \texttt{`ironing laundry'}, and \texttt{`renovating a room'}, and the entropy would be another measure to identify the complexity of the task captured by the compact graph representations.
% \AMC{No need to mention previous line in here, have it in appendix.} 

\section{Experimental Setup: Evaluating LLMs} \label{app:experimental_setup_additional}
As explained in the main paper, our overall evaluation relies on an MCQA-based Evaluation scheme that can generate $\sim 10^{17}$ commonsense queries for a single activity. 

\noindent \textbf{Commonsense Queries for Evaluation:} Note that though the proposed scheme is capable of generating enormous queries, 
we perform all the analysis on the dataset generated from 2k trajectories (leading to $\sim 20k$ commonsense queries) for each scenario. We freeze this dataset of ($\sim 20k$ commonsense queries per scenario) for easier replicability of the obtained results.

We provide details of the additional experiments to investigate the effectiveness
of these open-weight models below.

\noindent\textbf{In-Context Learning:} In recent years, LLMs have shown a surprising ability to capture the context via a few examples of the task provided in a prompt in the form of input-output examples \cite{dong2022survey}. The LLMs predict the next output conditioning on the previous examples. To quantify the performance of the created task, it becomes important to consider evaluating LLMs using in-context examples. We perform an evaluation of the created commonsense queries by considering 1-shot, 2-shot, and 5-shot experiments for the LLMs. Previously, a few of the works \cite{incontextfewshotlearners, robinson2023leveraging} have reported significant boosts in performance when provided with in-context examples for MCQA-based evaluation.

\noindent \textbf{Fine-Tuning:} We also consider the finetuning of 2 open-weight models (\texttt{phi-2} and \texttt{Llama-3}) over a small set of $1000$  queries from the created commonsense reasoning queries. We specifically choose 5 common scenarios and fine-tune the LLMs for an epoch. The fine-tuned scenarios include \texttt{planting a tree}, \texttt{going on a train}, \texttt{going grocery shopping}, \texttt{flying in an airplane}, and \texttt{riding on a bus}. We choose these five scenarios based on their generic nature, when compared to more specific scenarios like \texttt{`sewing a button'} or \texttt{`renoating a room'}.
% \AJ{better to provide a reason for choosing these models and scenarios}
% Table \ref{tab:llama3-finetuned}
%The fine-tuning took \AJ{mention the scenario names}. The fine-tuning took (\AJ{time for phi and llama to be added here}). 

\noindent \textbf{Generalization between Similar Scenarios:}  To assess if simple finetuning over a few scenarios helps the model learn the MCQA evaluation format, we consider evaluating the fine-tuned models over all the available scenarios. This also helps validate if there is a generalization between similar scenarios, i.e., learning a scenario helps improve the performance over other similar scenarios.

% \AMC{I think given the space limitation, we can briefly mention about various experiments and talk about the main findings and move most of the content to appendix; so that we get space to describe the localization experiment results.}

\section{Hyperparameters for Fine-Tuning Experiments} \label{sec-hyp_fine_tuning} 
We employed the following hyperparameters to fine-tune our models. We set the batch size to 4 and utilized gradient accumulation steps of 4. The models were trained for one epoch with a learning rate of $\mathbf{1e-5}$. A weight decay of 0.01 was applied. Flash attention \cite{dao2022flashattention} was enabled to enhance the training efficiency. The AdamW \cite{DBLP:journals/corr/abs-1711-05101} optimizer was used for updating the model weights.

\section{Additional Results and Empirical Findings} \label{app:additional_results}

\begin{figure}[t]
    \centering
     \includegraphics[width=\linewidth]{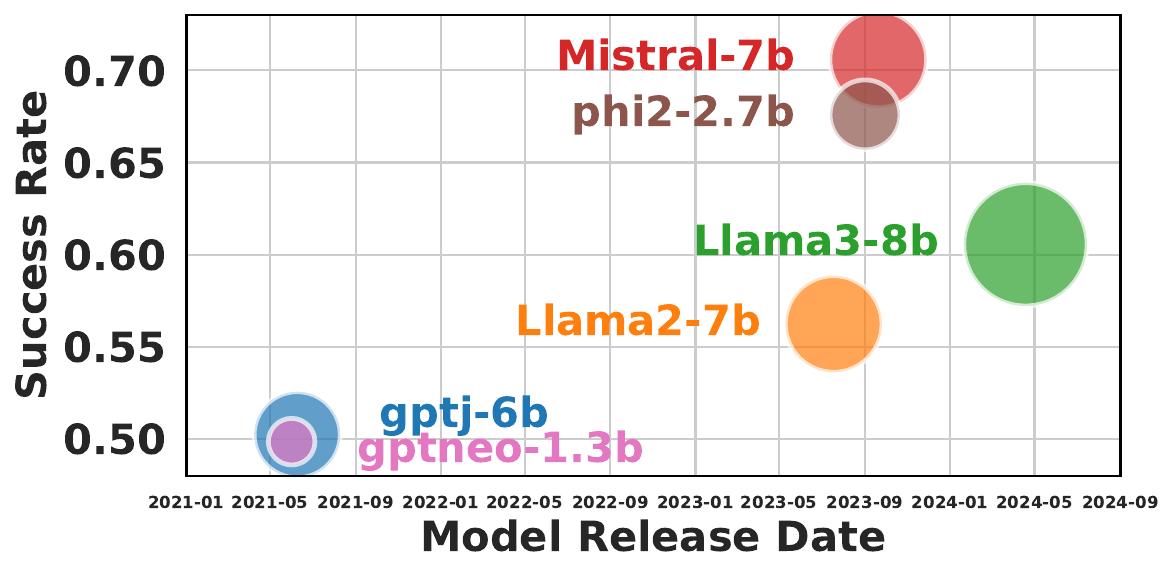}
    \caption{Comparing the success rates of the models on all the scenarios based on their release date and model size. The size of each circle is indicative of the number of parameters in the model. Here, we observe that \texttt{phi2} shows a considerable gap in performance when compared to \texttt{Llama} model series and is very close to \texttt{Mistral-7b} despite having less than half the number of parameters.}
    %% \AJ{2 different versions remaining}
    \label{fig:success-rate-vs-model-date}
\end{figure}

\noindent \textbf{Relation with Model Size:} 
% \AMC{I think this can be shortened and merged with the previous para} 
Fig. \ref{fig:success-rate-vs-model-date} underscores the success rate of a model compared to its size and release date. We observe a surprising trend in that \texttt{phi2-2.7b} can outperform the Llama series of models despite its smaller size. Through Fig. \ref{fig:success-rate-vs-model-date}, we observe that this performance rise could be attributed to the release dates of the models and the availability of pre-training datasets.

% \noindent \textbf{Performance across various Scenarios:} We further explore model performance in Fig. \ref{fig:task-completion-vs-success-rate-all-models-all-scenarios} by comparing the success rates of the models against the task completion percentage. 
% The task completion percentage is calculated based on the current event step with respect to the total steps in the sampled trajectory.
% We observe that all models perform well for smaller task completion percentage; however, as the task progresses, all the models show a dip in success rates. This could possibly be attributed to either the long context of all the previous actions or the task's complexity as it progresses. In general, LLMs are expected to perform well with more context about the task (note the context length here does not increase by a significant margin). However, in this case, as the task progresses, the number of valid options increases with more variability, increasing the complexity of the commonsense queries. 

% This is also evident from the compact graphs that each scenario has a single starting point, but as we progress in the task, the number of options rises significantly, increasing the complexity of the commonsense queries. 

\noindent \textbf{Relation with Task completion percentage:} 
% \AMC{I think we can mention about it briefly in here and move the rest to appendix.}
The MCQA formulation of the commonsense knowledge about the activities is framed using the steps/events involved in the activity, where 
a subpart of the trajectory (with length $m$) is considered by taking a split at a step $n \in \{1, m\}$ and using steps $e_1, e_2, \ldots e_{n-1}$ as a part of a commonsense reasoning question and $e_n$ as the correct choice for the question. 
The task completion percentage is calculated based on the current event step ($n$) with respect to the total steps ($m$) in the sampled trajectory.
% We monitor the task completion percentage for a particular commonsense query via the location of node $n$ in the compact graph. 
More task completion percentage means more context of a particular task, i.e. the query contains more number of steps. 

\begin{figure}[t]
    \centering
    \includegraphics[width=0.8\linewidth]{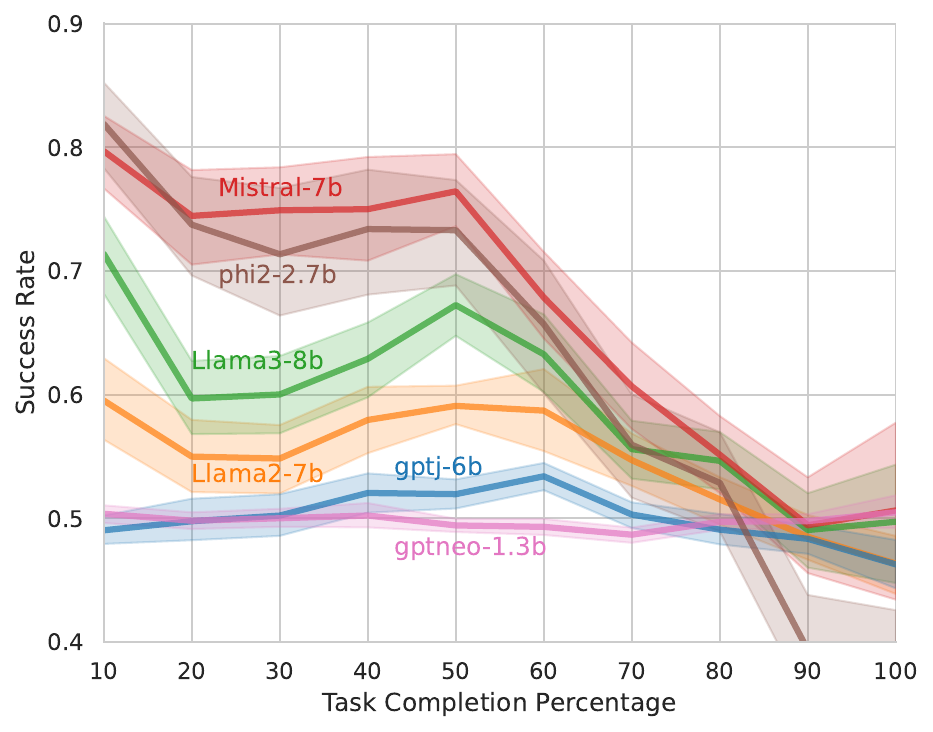}
    \caption{Comparing the success rates of the models across task completion percentage. The error bands show +1 and -1 standard deviations across scenarios and in-context shots.} %\AJ{same plot for multiple shots can be done and moved to the appendix; we can also compare scenarios in a bigger plot if needed, which would also be interesting to highlight in the appendix (task completion percentage with scenarios)}}
    \label{fig:task-completion-vs-success-rate-all-models-all-scenarios}
    %%\vspace{-6mm}
\end{figure}

We investigate model performance in Fig. \ref{fig:task-completion-vs-success-rate-all-models-all-scenarios} by comparing the success rates of the models against the task completion percentage.
We observe that all models perform well for smaller task completion percentage; however, as the task progresses, all the models show a dip in success rates. This could possibly be attributed to either the long context of all the previous actions or the task's complexity as it progresses. In general, LLMs are expected to perform well with more context about the task (note the context length here does not increase by a significant margin). However, in this case, as the task progresses, the number of valid options increases with more variability, increasing the complexity of the commonsense queries. 

\begin{figure*}
    \centering
    \includegraphics[scale=0.21]{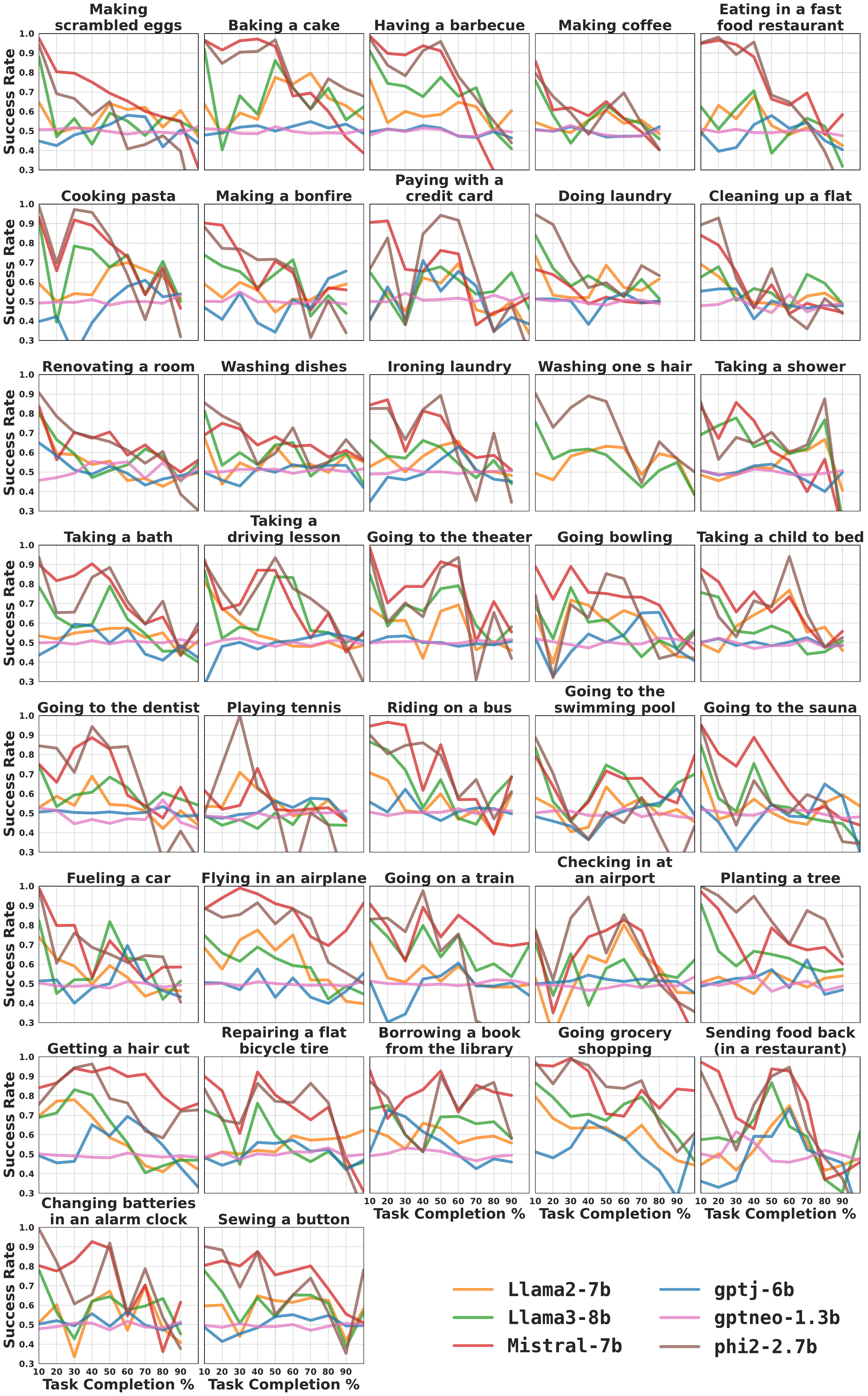}
    \caption{Comparing success rates of the presented 6 models across each scenario and task completion percentages in a 5-shot setting. Here we see that for many scenarios \texttt{phi2-2.7b} and \texttt{Mistral-7b} show similar success rates. All the models have a high success rate earlier in each task, however as the task progresses the models show a drop in success. \texttt{gptj-6b} and \texttt{gptneo-1.3b} show almost random success ($\approx 50\%$) on each task. 
    % \AJ{another version of this plot is pending, can keep it in the appendix} \AJ{the font can be made bold here with slight increase in font size for better visibility.}
    }
    \label{fig:task-completion-vs-success-rate-5-shot-for-each_script-world-scenarios}
\end{figure*}

To further explore how the performance varies with task completion percentage for different scenarios, we compare the performance across all the scenarios. 
Fig. \ref{fig:task-completion-vs-success-rate-5-shot-for-each_script-world-scenarios} shows the success rate of each model across task completion percentages for each scenario. We observe a similar trend and notice that all models perform well initially but show a decline in performance thereafter. 
% \texttt{Mistral-7b} shows remarkable success rates in \texttt{Getting a haricut} across all task completion steps. 

We observed a few interesting trends when inspecting them across similar scenarios. For the scenarios that contain relationships with food, for example, in scenarios like \texttt{Making scrambled eggs}, \texttt{Baking a cake}, \texttt{Having a barbecue}, \texttt{Making coffee}, etc., the \texttt{Mistral-7b} shows a significant drop in the performance towards the end, highlighting the role of context in making the task more detailed and difficult to reason about. Moreover, we also find an interesting trend where the scenarios contain some movement, e.g., \texttt{Taking a driving lesson}, \texttt{Going to the theatre}, \texttt{Going bowling}, \texttt{Taking a child to bed}, \texttt{Going to the dentist}, \texttt{Riding on a bus}, \texttt{Flying in an airplane}, and \texttt{Checking in at the airport}; \texttt{Mistral-7b} and \texttt{phi2-2.7b} show improvements in success rates at the middle sections of the task, making the context more important for such scenarios.

% \noindent \textbf{Task complexity vs. Performance:} \AJ{can include the node degree or total number of trajectories from Table 1 and plot it with performance across multiple scenarios, it can be a scatter plot highlighting the observed trend.}

\begin{figure}[t]
    \centering
    \includegraphics[width=\linewidth]{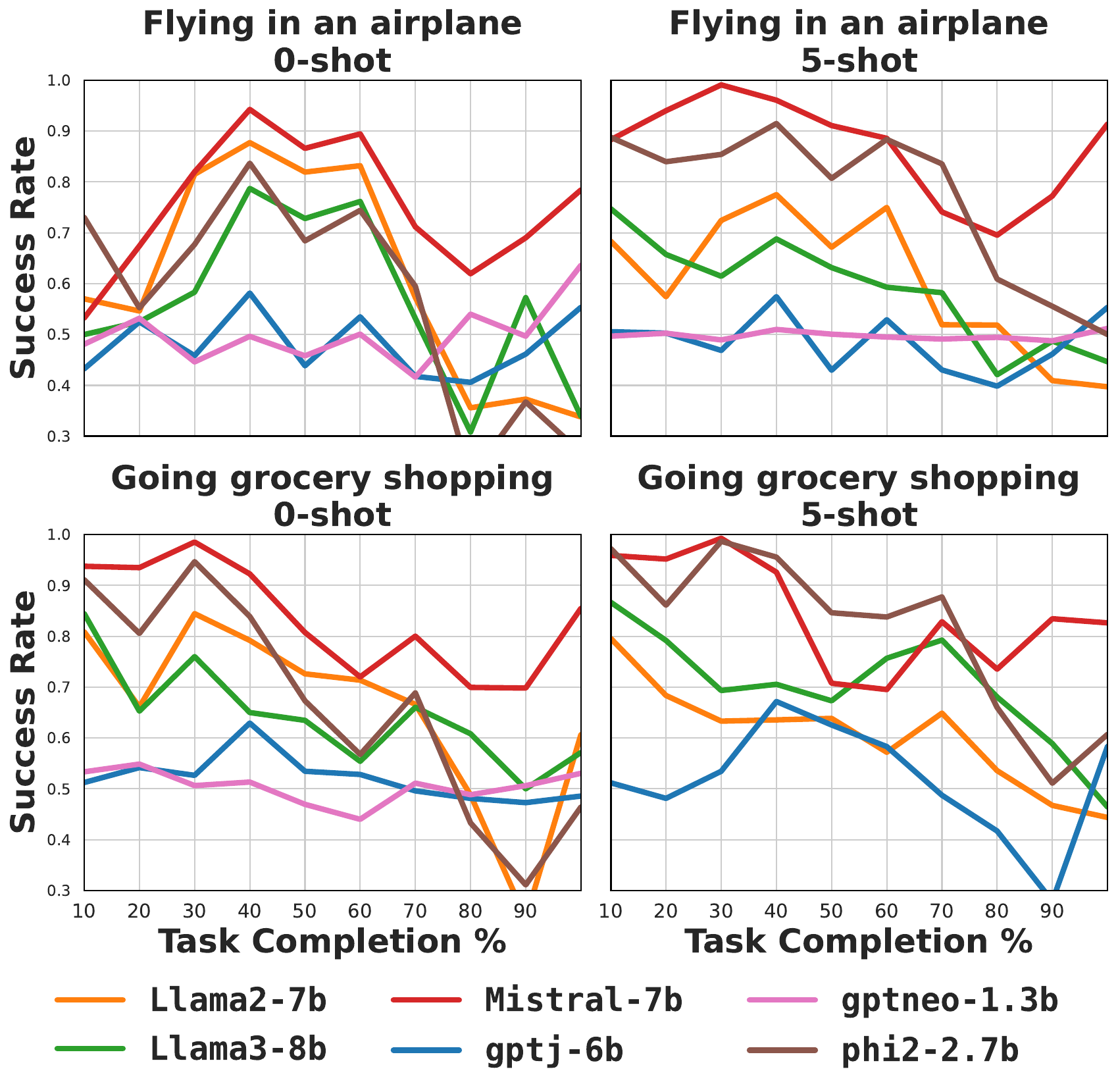}
    \caption{\texttt{Flying an airplane} and \texttt{Going grocery shopping} and show considerable improvement in \texttt{phi2-2.7b} and \texttt{Mistral-7b} when going from 0-shot to 5-shot.}
    \label{fig:task-completion-vs-success-rate-0-5-shot-for-going-grocery-shopping-and-flying-in-an-airplane}
\end{figure}

\noindent \textbf{Improvements with In-Context Learning Examples:} 
% \AMC{I think this can be shortened and move major part to appendix} 
Fig. \ref{fig:task-completion-vs-success-rate-0-5-shot-for-going-grocery-shopping-and-flying-in-an-airplane} shows the improvement of the models from zero-shot to five-shot settings, especially at the initial steps. \texttt{Mistral-7b}, \texttt{phi2-2.7b}, \texttt{Llama3-8b}, and \texttt{gptj-6b} show performance improvements in the \texttt{Going Grocery Shopping} scenario (holding the highest scores in \texttt{Mistral-7b}). However, \texttt{Llama2-7b} shows performance degradation when going from 0-shot to 5-shot 
% \AM{some possible reason for this}
. A similar trend is observed in \texttt{Flying in an Airplane} scenario, with \texttt{Mistral-7b} and \texttt{phi2-2.7b} showing performance improvements while Llama models show a degradation in performance 
% \AM{some possible reason for this}
. 

\begin{figure}[t]
    \centering
     \includegraphics[width=0.8\linewidth]{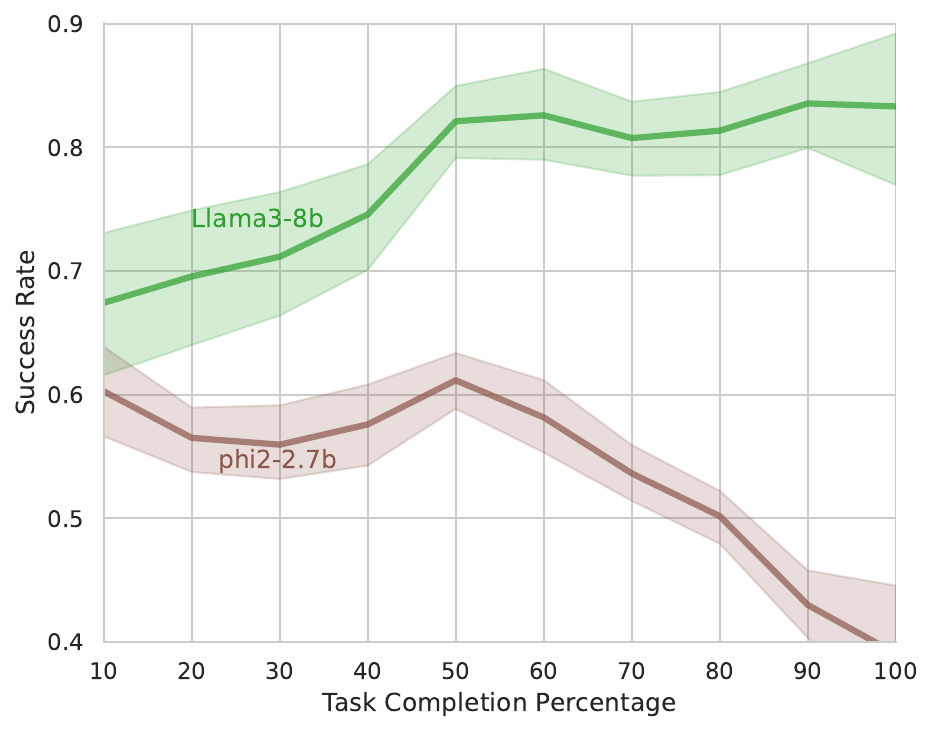}
    \caption{Success rate of the models after fine-tuning it on the MCQA dataset. The error bands show +1 and -1 standard deviation across scenarios.}
    \label{fig:task-completion-vs-success-rate-all-models-all-scenarios-finetuned}
    %%\vspace{-6mm}
\end{figure}

\begin{figure*}
    \centering
    \includegraphics[width=0.85\textwidth]{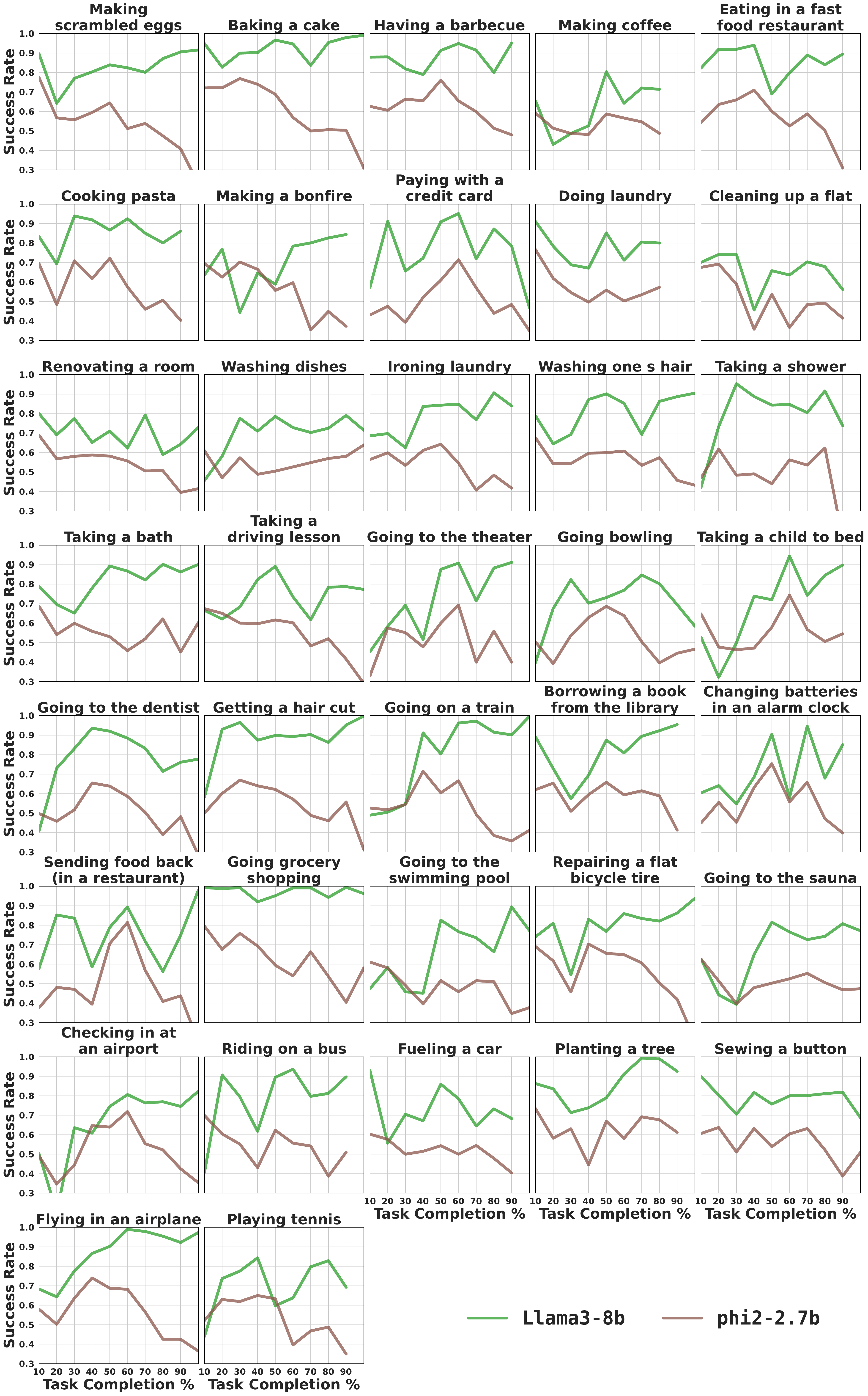}
    \caption{Task completion \% vs success rate of all models on each scenario for fine-tuned \texttt{Llama3-8b} and \texttt{phi2-2.7b}}
    \label{fig:task-completion-vs-success-rate-all-shot-for-each-script-world-scenarios-finetuned}
\end{figure*}

\noindent \textbf{Improvements with Finetuning:} 
% \AMC{I think this can be shortened and move major part to appendix} 
Table \ref{tab:llama3-finetuned} and  Fig. \ref{fig:task-completion-vs-success-rate-all-models-all-scenarios-finetuned} highlights the improvement of \texttt{Llama3} across all scenarios and task completion status upon fine-tuning. We observe that after fine-tuning \texttt{Llama3}, it has a rise in success rate across time steps and also outperform \texttt{Mistral-7B} when prompted with in-context examples. Fig. \ref{fig:task-completion-vs-success-rate-all-shot-for-each-script-world-scenarios-finetuned} dives deeper into the fine-tuned models across time steps for each scenario. We also observe the same rising trend, suggesting that the model generalize upon fine-tuning. However, we observe a decrease in performance of \texttt{phi2-2.7b} in general and across time steps. We observe the same trend in Fig. \ref{fig:task-completion-vs-success-rate-all-shot-for-each-script-world-scenarios-finetuned} for \texttt{phi2-2.7b} in all scenarios.

\noindent \textbf{Generalization across Multiple Scenarios:}
We fine-tuned \texttt{Llama3-8b} and \texttt{phi2-2.7b} on the trajectories from 
\texttt{Going grocery shopping} 
and evaluated the models on all the scenarios. Fig. \ref{fig:task-completion-vs-success-rate-all-models-all-scenarios-finetuned} highlights that \texttt{Llama3-8b} generalizes to all the scenarios, especially across time steps. However, we see a drop in the performance of \texttt{phi2-2.7b} in general and across time steps, pointing towards low generalization capability of smaller models.
% \AM{some possible reason for this}.

% \AMC{make the font of all model names consistent in text TeleType}
%\AJ{some conclusive remarks about the observed trends/findings will go here}

\begin{table}[t]
\centering
\small
\renewcommand{\arraystretch}{1}
\begin{tabular}{@{}ccccc@{}}
\toprule
\textit{\textbf{Language Model}} & \textbf{0-shot} & \textbf{1-shot} & \textbf{2-shot} & \textbf{5-shot} \\ \midrule
gpt-j-6B                         & 50.19                      & 50.14                      & 50.59                      & 50.05                      \\
gpt-neo-1.3B                     & 50.07                      & 49.58                      & 49.86                      & 50.26                      \\
Llama-2-7b-chat-hf               & 55.67                      & 54.63                      & 56.11                      & 56.59                      \\
Mistral-7B-v0.1                  & 66.76                      & 67.61                      & 70.24                      & 71.13                      \\ \midrule
\textbf{Average}             & \textbf{55.99}             & \textbf{55.78}             & \textbf{57.04}             & \textbf{57.26}             \\ \bottomrule
\end{tabular}
\caption{Average performance for In-context learning experiments over multiple open-weight models.}
%%\vspace{-4mm}
\end{table}

\begin{figure}
    \centering
    \includegraphics[width=\linewidth]{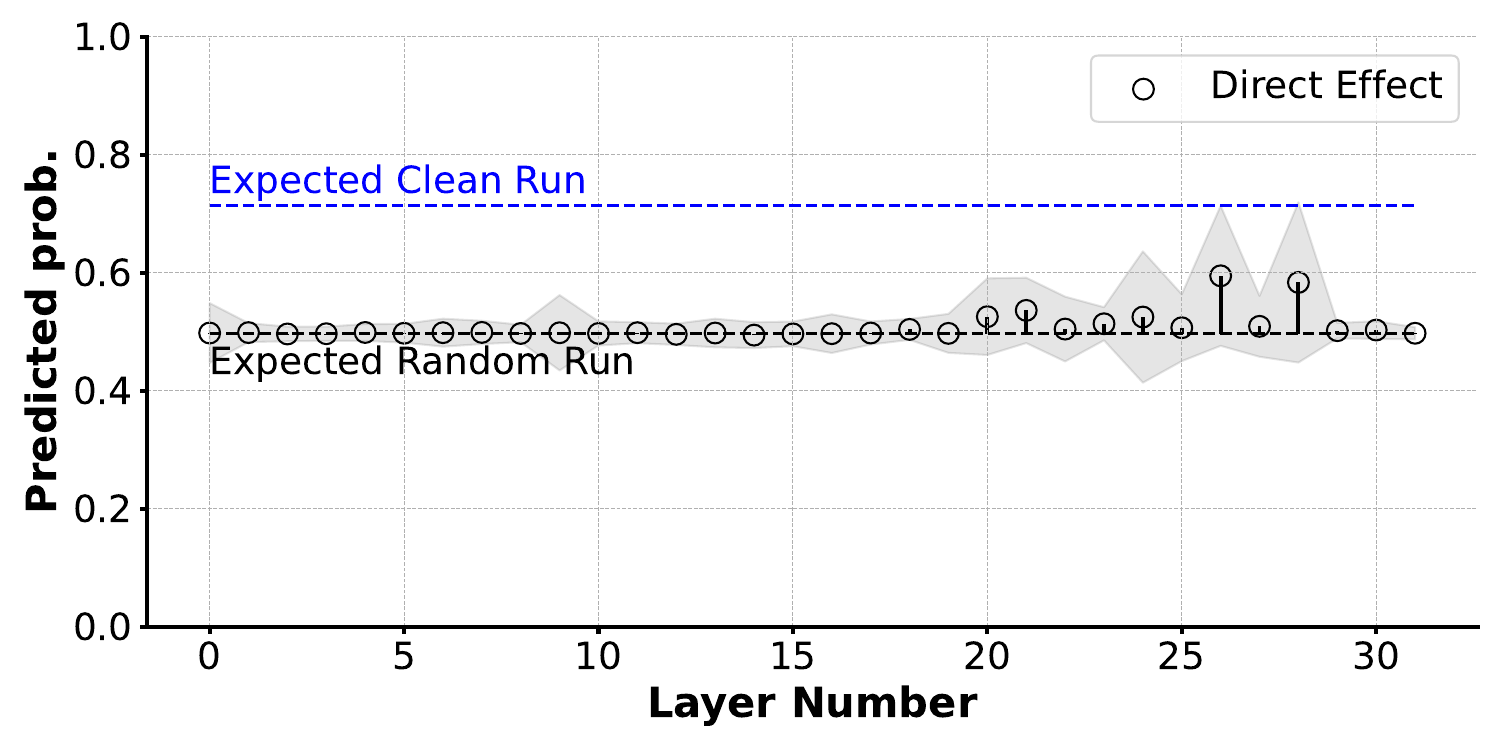}
    \caption{ The figure shows the direct effect of path patching from the clean run to the random run (`\textit{going bowling}'). The peaks/deviations are less decisive than \ref{fig:decision-localization}, highlighting the effectiveness of using the proposed conjugate prompts.
    % leading to deviations starting at layer 20 and increased signal strength at layer 26, highlighting the role of particular layers in commonsense reasoning.
    }
    \label{fig:decision-localization-random}
    \vspace{-5mm}
\end{figure}

% \begin{figure}
%     \centering
%     \includegraphics[width=0.45\linewidth]{./images/success-rate_vs_model-size.pdf}
%     \caption{Comparing the success rates of the models on all the scenarios based on their size. The size of each circle is indicative of the number of parameters in the model. Here we observe that \texttt{phi2} shows a considerable gap in performance when compared to Llama model series and is very close to \texttt{Mistral-7b} despite having less than half the number of parameters.
%     \AJ{3 different versions remaining}
%     }
%     \label{fig:success-rate-vs-model-size}
% \end{figure}

\begin{table*}[t]
    \centering
    \small
    \renewcommand{\arraystretch}{1.2}
    \resizebox{0.95\linewidth}{!}
    {%
    \begin{tabular}{@{}ccccccc@{}}
    \toprule
    \textbf{Activity}                       & \textbf{gpt-neo-1.3B}         & \textbf{phi-2}            & \textbf{gpt-j-6B}             & \textbf{Llama-2-7b}   & \textbf{Mistral-7B-v0.1}&\textbf{Llama-3} \\
    \midrule
    \texttt{baking a cake }& 48.86 & 73.78 & 44.43 &69.28 & \textbf{77.84} & 63.11                \\ 
    \texttt{borrowing a book from the library} & 49.37 & 60.76 & 52.45 & 61.86 & \textbf{75.08} & 55.41                \\ 
    \texttt{changing batteries in an alarm clock} & 50.02 & \textbf{62.79} & 50.65 & 51.51 & 60.53 & 51.19                \\ 
    \texttt{checking in at an airport} & 48.77 & 55.65 & 48.60 &53.74 & \textbf{58.16} & 48.76                \\ 
    \texttt{cleaning up a flat} & 49.23 & 57.93 & 51.29 & 50.64 & \textbf{59.58} & 51.84               \\ 
    \texttt{cooking pasta} & 51.88 & 67.03 & 49.68 & 60.56 & \textbf{70.20} & 60.31               \\ 
    \texttt{doing laundry} & 49.43 & 68.54 & 50.27 & 60.84 & \textbf{73.26} & 56.46            \\ 
    \texttt{eating in a fast food restaurant} & 49.91 & 63.00 & 52.19 & 50.62 & \textbf{71.74} & 53.64                \\ 
    \texttt{flying in an airplane} & 49.83 & 59.43 & 48.30 & 62.40 & \textbf{74.14} & 57.93                \\ 
    \texttt{fueling a car} & 51.05 & 59.78 & 51.30 & 45.54 & \textbf{64.44} & 50.79                \\ 
    \texttt{getting a hair cut} & 49.91 & 56.81 & 48.53 & 54.72 & \textbf{73.75} & 50.58                \\ 
    \texttt{going bowling} & 49.87 & 58.05 & 50.59 & 53.14 & \textbf{61.15} & 49.29                \\ 
    \texttt{going grocery shopping}               & 50.84                & 70.96              & 53.12              & 67.36              & \textbf{84.96}           & 66.77              \\
    \texttt{going on a train} & 50.16 & 55.00 & 52.03 & 58.10 & \textbf{69.81} & 54.70               \\ 
    \texttt{going to the dentist} & 50.29 & 54.28 & 51.56 & 52.10 & \textbf{66.58} & 52.57                \\ 
    \texttt{going to the sauna} & 50.47 & 53.17 & 49.91 & 50.70 & \textbf{57.50} & 50.83                 \\ 
    \texttt{going to the swimming pool} & 47.90 & 54.34 & 49.51 & 46.93 & \textbf{57.16} & 46.72                \\ 
    \texttt{going to the theater} & 48.49 & 52.76 & 51.94 & 48.54 & \textbf{61.35} & 47.75                \\ 
    \texttt{having a barbecue} & 48.41 & \textbf{77.19} & 52.33 & 60.31 & 76.30 & 57.97                 \\ 
    \texttt{ironing laundry} & 51.69 & 61.57 & 48.65 & 57.88 & \textbf{65.04} & 51.81                 \\
    \texttt{making a bonfire} & 51.17 & \textbf{65.22} & 49.22 & 51.33 & 64.29 & 48.00                \\ 
    \texttt{making coffee} & 51.62 & 57.51 & 49.44 & 51.04 & \textbf{59.95} & 49.91              \\ 
    \texttt{making scrambled eggs} & 51.49 & \textbf{66.08} & 48.64 & 56.46 & 65.99 & 57.84                \\ 
    \texttt{paying with a credit card}            & 49.15                & 38.92               & 50.49                & 49.07                & \textbf{50.95}           & 49.84               \\
    \texttt{planting a tree} & 49.60 & 71.25 & 49.18 & 63.59 & \textbf{73.19} & 60.27              \\ 
    \texttt{playing tennis} & 48.65 & 56.09 & 50.87 & 47.67 & \textbf{64.96} & 50.18               \\ 
    \texttt{renovating a room} & 47.09 & 60.92 & 51.38 & 52.45 & \textbf{63.49} & 51.06                \\ 
    \texttt{repairing a flat bicycle tire} & 50.38 & \textbf{71.32} & 50.26 & 59.05 & 69.59 & 55.59                \\ 
    \texttt{riding on a bus} & 48.04 & 61.99 & 53.31 & 58.39 & \textbf{71.99} & 54.98               \\ 
    \texttt{sending food back (in a restaurant)} & 53.98 & 49.23 & 48.37 & 50.84 & \textbf{63.69} & 51.15               \\ 
    \texttt{sewing a button} & 51.95 & 63.06 & 48.53 & 54.28 & \textbf{66.68} & 52.70              \\ 
    \texttt{taking a bath} & 49.91 & 59.31 & 49.54 & 55.32 & \textbf{69.52} & 52.74               \\ 
    \texttt{taking a child to bed} & 51.56 & 60.21 & 49.55 & 54.69 & \textbf{68.74} & 54.25               \\ 
    \texttt{taking a driving lesson} & 49.67 & \textbf{63.97} & 51.32 & 59.64 & 63.27 & 54.28              \\ 
    \texttt{taking a shower} & 49.94 & 49.93 & 50.17 & 57.77 & \textbf{68.33} & 55.61               \\ 
    \texttt{washing dishes} & 49.99 & \textbf{62.32} & 50.73 & 52.32 & 60.07 & 50.26              \\ 
    \texttt{washing one s hair} & 48.63 & 64.52 & 50.15 & 53.40 & \textbf{66.75} & 57.12                \\ 
    \midrule
    \textbf{Average Performance} & 50.01 & 60.67 & 50.23 & 55.27 & \textbf{66.76} & 53.63 \\
    \bottomrule
    \end{tabular}
    }
    \caption{Success Rate (\%) of various open-weight LLMs over the created commonsense queries for \scenariocount real-world activities. Overall, we find Mistral-7B-v0.1 performing best over the maximum number of scenarios, highlighting better commonsense reasoning abilities when compared to other open-weight models. We also observe that phi-2, with a surprisingly lower number of parameters, outperforms models with more number of parameters.}
    \label{tab:zero-shot-results}
    % \vspace{-2mm}
    \end{table*}

\begin{figure*}
    \centering
    \includegraphics[width=0.85\textwidth]{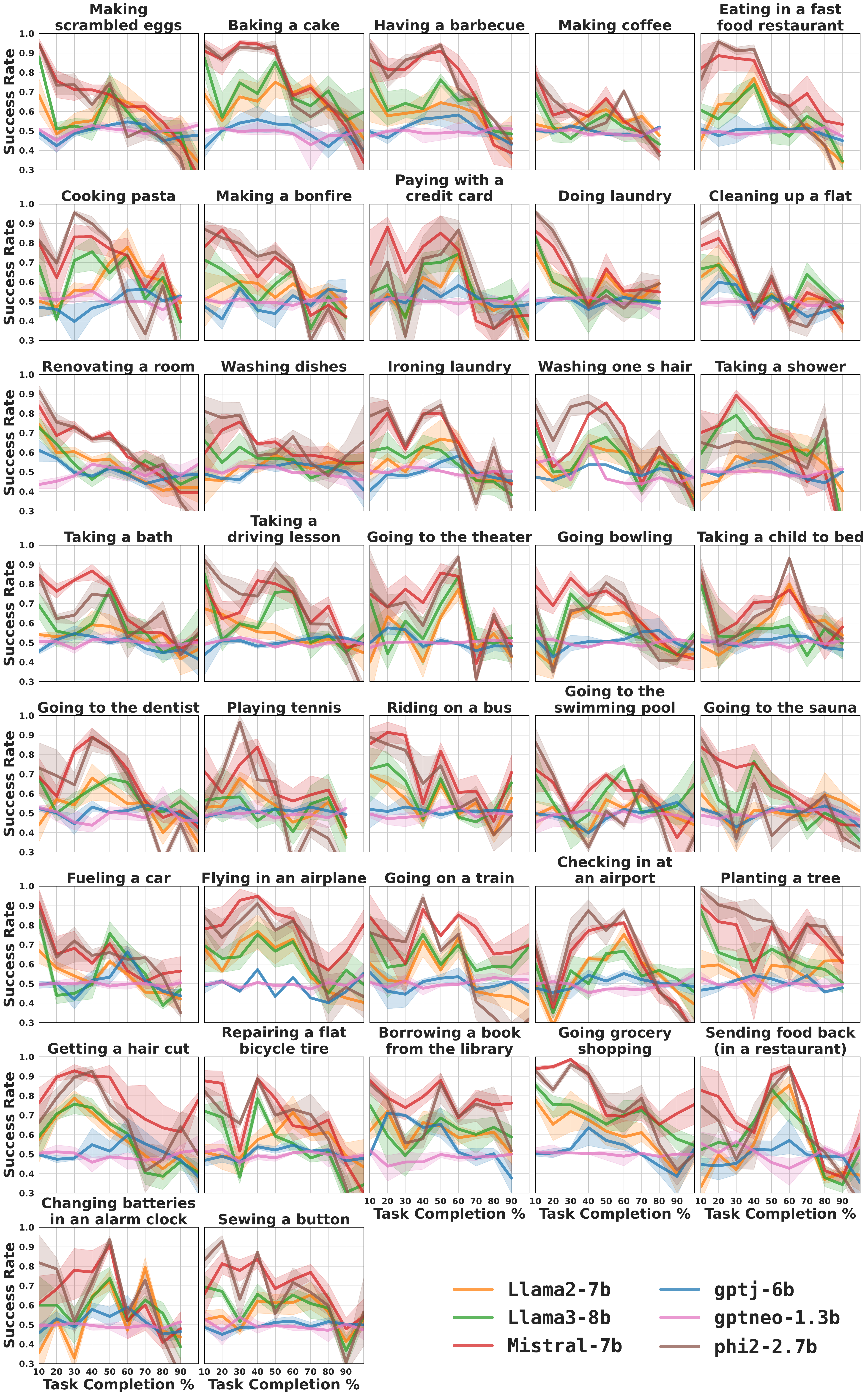}
    \caption{Task completion \% vs success rate of all models on each scenario averaged over all number of in-context examples, i.e. n-shots}
    \label{fig:task-completion-vs-success-rate-all-shot-for-each-script-world-scenarios}
\end{figure*}

\begin{table*}[t]
    \centering
    \small
    \setlength\tabcolsep{4pt}
    \renewcommand{\arraystretch}{1.2}
    \resizebox{\linewidth}{!}{%
    \begin{tabular}{@{}cccccc@{}}
    \toprule
    \textbf{Scenario} & \texttt{planting a tree} & \texttt{going on a train} & \texttt{going grocery shopping} & \texttt{flying in an airplane} & \texttt{riding on a bus} \\
     \midrule

\texttt{baking a cake} & 91.58 (\textcolor{blue}{$\uparrow$} \textcolor{blue}{28.47\%}) & 94.09 (\textcolor{blue}{$\uparrow$} \textcolor{blue}{30.98\%}) & 92.59 (\textcolor{blue}{$\uparrow$} \textcolor{blue}{29.48\%}) & 93.21 (\textcolor{blue}{$\uparrow$} \textcolor{blue}{30.10\%}) & 82.23 (\textcolor{blue}{$\uparrow$} \textcolor{blue}{19.12\%})  \\

\texttt{borrowing a book from the library} & 82.26 (\textcolor{blue}{$\uparrow$} \textcolor{blue}{26.85\%}) & 86.55 (\textcolor{blue}{$\uparrow$} \textcolor{blue}{31.14\%}) & 86.44 (\textcolor{blue}{$\uparrow$} \textcolor{blue}{31.03\%}) & 84.49 (\textcolor{blue}{$\uparrow$} \textcolor{blue}{29.08\%}) & 81.14 (\textcolor{blue}{$\uparrow$} \textcolor{blue}{25.73\%})  \\

\texttt{changing batteries in an alarm clock} & 81.17 (\textcolor{blue}{$\uparrow$} \textcolor{blue}{29.98\%}) & 79.09 (\textcolor{blue}{$\uparrow$} \textcolor{blue}{27.90\%}) & 74.13 (\textcolor{blue}{$\uparrow$} \textcolor{blue}{22.94\%}) & 74.86 (\textcolor{blue}{$\uparrow$} \textcolor{blue}{23.67\%}) & 75.23 (\textcolor{blue}{$\uparrow$} \textcolor{blue}{24.04\%})  \\

\texttt{checking in at an airport} & 62.29 (\textcolor{blue}{$\uparrow$} \textcolor{blue}{13.53\%}) & 67.31 (\textcolor{blue}{$\uparrow$} \textcolor{blue}{18.55\%}) & 67.07 (\textcolor{blue}{$\uparrow$} \textcolor{blue}{18.31\%}) & 68.58 (\textcolor{blue}{$\uparrow$} \textcolor{blue}{19.82\%}) & 61.22 (\textcolor{blue}{$\uparrow$} \textcolor{blue}{12.46\%})  \\

\texttt{cleaning up a flat} & 63.55 (\textcolor{blue}{$\uparrow$} \textcolor{blue}{11.71\%}) & 63.36 (\textcolor{blue}{$\uparrow$} \textcolor{blue}{11.52\%}) & 65.41 (\textcolor{blue}{$\uparrow$} \textcolor{blue}{13.57\%}) & 65.23 (\textcolor{blue}{$\uparrow$} \textcolor{blue}{13.39\%}) & 65.38 (\textcolor{blue}{$\uparrow$} \textcolor{blue}{13.54\%})  \\

\texttt{cooking pasta} & 84.45 (\textcolor{blue}{$\uparrow$} \textcolor{blue}{24.14\%}) & 83.15 (\textcolor{blue}{$\uparrow$} \textcolor{blue}{22.84\%}) & 86.65 (\textcolor{blue}{$\uparrow$} \textcolor{blue}{26.34\%}) & 83.62 (\textcolor{blue}{$\uparrow$} \textcolor{blue}{23.31\%}) & 82.37 (\textcolor{blue}{$\uparrow$} \textcolor{blue}{22.06\%})  \\

\texttt{doing laundry} & 76.06 (\textcolor{blue}{$\uparrow$} \textcolor{blue}{19.60\%}) & 80.14 (\textcolor{blue}{$\uparrow$} \textcolor{blue}{23.68\%}) & 79.08 (\textcolor{blue}{$\uparrow$} \textcolor{blue}{22.62\%}) & 79.29 (\textcolor{blue}{$\uparrow$} \textcolor{blue}{22.83\%}) & 76.09 (\textcolor{blue}{$\uparrow$} \textcolor{blue}{19.63\%})  \\

\texttt{eating in a fast food restaurant} & 78.84 (\textcolor{blue}{$\uparrow$} \textcolor{blue}{25.19\%}) & 84.95 (\textcolor{blue}{$\uparrow$} \textcolor{blue}{31.30\%}) & 85.69 (\textcolor{blue}{$\uparrow$} \textcolor{blue}{32.04\%}) & 87.35 (\textcolor{blue}{$\uparrow$} \textcolor{blue}{33.70\%}) & 80.45 (\textcolor{blue}{$\uparrow$} \textcolor{blue}{26.80\%})  \\

\rowcolor{purple!15}\texttt{flying in an airplane} & 86.81 (\textcolor{blue}{$\uparrow$} \textcolor{blue}{28.88\%}) & 90.15 (\textcolor{blue}{$\uparrow$} \textcolor{blue}{32.22\%}) & 84.44 (\textcolor{blue}{$\uparrow$} \textcolor{blue}{26.51\%}) & 95.71 (\textcolor{blue}{$\uparrow$} \textcolor{blue}{37.78\%}) & 77.50 (\textcolor{blue}{$\uparrow$} \textcolor{blue}{19.57\%})  \\

\texttt{fueling a car} & 73.19 (\textcolor{blue}{$\uparrow$} \textcolor{blue}{22.40\%}) & 73.07 (\textcolor{blue}{$\uparrow$} \textcolor{blue}{22.28\%}) & 73.44 (\textcolor{blue}{$\uparrow$} \textcolor{blue}{22.65\%}) & 74.48 (\textcolor{blue}{$\uparrow$} \textcolor{blue}{23.69\%}) & 73.57 (\textcolor{blue}{$\uparrow$} \textcolor{blue}{22.78\%})  \\

\texttt{getting a hair cut} & 84.83 (\textcolor{blue}{$\uparrow$} \textcolor{blue}{34.25\%}) & 85.45 (\textcolor{blue}{$\uparrow$} \textcolor{blue}{34.87\%}) & 85.60 (\textcolor{blue}{$\uparrow$} \textcolor{blue}{35.02\%}) & 87.89 (\textcolor{blue}{$\uparrow$} \textcolor{blue}{37.31\%}) & 79.17 (\textcolor{blue}{$\uparrow$} \textcolor{blue}{28.59\%})  \\

\texttt{going bowling} & 66.00 (\textcolor{blue}{$\uparrow$} \textcolor{blue}{16.71\%}) & 69.63 (\textcolor{blue}{$\uparrow$} \textcolor{blue}{20.34\%}) & 67.37 (\textcolor{blue}{$\uparrow$} \textcolor{blue}{18.08\%}) & 69.89 (\textcolor{blue}{$\uparrow$} \textcolor{blue}{20.60\%}) & 68.66 (\textcolor{blue}{$\uparrow$} \textcolor{blue}{19.37\%})  \\

\rowcolor{orange!15}\texttt{going grocery shopping} & 90.69 (\textcolor{blue}{$\uparrow$} \textcolor{blue}{23.92\%}) & 92.69 (\textcolor{blue}{$\uparrow$} \textcolor{blue}{25.92\%}) & 96.82 (\textcolor{blue}{$\uparrow$} \textcolor{blue}{30.05\%}) & 94.17 (\textcolor{blue}{$\uparrow$} \textcolor{blue}{27.40\%}) & 87.66 (\textcolor{blue}{$\uparrow$} \textcolor{blue}{20.89\%})  \\

\rowcolor{navyblue!15}\texttt{going on a train} & 79.66 (\textcolor{blue}{$\uparrow$} \textcolor{blue}{24.96\%}) & 93.89 (\textcolor{blue}{$\uparrow$} \textcolor{blue}{39.19\%}) & 78.45 (\textcolor{blue}{$\uparrow$} \textcolor{blue}{23.75\%}) & 86.35 (\textcolor{blue}{$\uparrow$} \textcolor{blue}{31.65\%}) & 75.09 (\textcolor{blue}{$\uparrow$} \textcolor{blue}{20.39\%})  \\

\texttt{going to the dentist} & 68.95 (\textcolor{blue}{$\uparrow$} \textcolor{blue}{16.38\%}) & 78.18 (\textcolor{blue}{$\uparrow$} \textcolor{blue}{25.61\%}) & 77.02 (\textcolor{blue}{$\uparrow$} \textcolor{blue}{24.45\%}) & 79.14 (\textcolor{blue}{$\uparrow$} \textcolor{blue}{26.57\%}) & 71.23 (\textcolor{blue}{$\uparrow$} \textcolor{blue}{18.66\%})  \\

\texttt{going to the sauna} & 66.59 (\textcolor{blue}{$\uparrow$} \textcolor{blue}{15.76\%}) & 72.87 (\textcolor{blue}{$\uparrow$} \textcolor{blue}{22.04\%}) & 66.80 (\textcolor{blue}{$\uparrow$} \textcolor{blue}{15.97\%}) & 70.92 (\textcolor{blue}{$\uparrow$} \textcolor{blue}{20.09\%}) & 65.05 (\textcolor{blue}{$\uparrow$} \textcolor{blue}{14.22\%})  \\

\texttt{going to the swimming pool} & 66.51 (\textcolor{blue}{$\uparrow$} \textcolor{blue}{19.78\%}) & 69.39 (\textcolor{blue}{$\uparrow$} \textcolor{blue}{22.66\%}) & 63.75 (\textcolor{blue}{$\uparrow$} \textcolor{blue}{17.02\%}) & 71.40 (\textcolor{blue}{$\uparrow$} \textcolor{blue}{24.67\%}) & 62.85 (\textcolor{blue}{$\uparrow$} \textcolor{blue}{16.12\%})  \\

\texttt{going to the theater} & 71.62 (\textcolor{blue}{$\uparrow$} \textcolor{blue}{23.90\%}) & 74.62 (\textcolor{blue}{$\uparrow$} \textcolor{blue}{26.90\%}) & 71.81 (\textcolor{blue}{$\uparrow$} \textcolor{blue}{24.09\%}) & 80.23 (\textcolor{blue}{$\uparrow$} \textcolor{blue}{32.51\%}) & 68.64 (\textcolor{blue}{$\uparrow$} \textcolor{blue}{20.92\%})  \\

\texttt{having a barbecue} & 87.32 (\textcolor{blue}{$\uparrow$} \textcolor{blue}{29.34\%}) & 86.63 (\textcolor{blue}{$\uparrow$} \textcolor{blue}{28.65\%}) & 87.92 (\textcolor{blue}{$\uparrow$} \textcolor{blue}{29.94\%}) & 87.47 (\textcolor{blue}{$\uparrow$} \textcolor{blue}{29.49\%}) & 80.32 (\textcolor{blue}{$\uparrow$} \textcolor{blue}{22.34\%})  \\

\texttt{ironing laundry} & 73.86 (\textcolor{blue}{$\uparrow$} \textcolor{blue}{22.05\%}) & 78.18 (\textcolor{blue}{$\uparrow$} \textcolor{blue}{26.37\%}) & 78.57 (\textcolor{blue}{$\uparrow$} \textcolor{blue}{26.76\%}) & 78.68 (\textcolor{blue}{$\uparrow$} \textcolor{blue}{26.87\%}) & 77.30 (\textcolor{blue}{$\uparrow$} \textcolor{blue}{25.49\%})  \\

\texttt{making a bonfire} & 77.72 (\textcolor{blue}{$\uparrow$} \textcolor{blue}{29.72\%}) & 77.87 (\textcolor{blue}{$\uparrow$} \textcolor{blue}{29.87\%}) & 72.06 (\textcolor{blue}{$\uparrow$} \textcolor{blue}{24.06\%}) & 75.50 (\textcolor{blue}{$\uparrow$} \textcolor{blue}{27.50\%}) & 67.50 (\textcolor{blue}{$\uparrow$} \textcolor{blue}{19.50\%})  \\

\texttt{making coffee} & 70.11 (\textcolor{blue}{$\uparrow$} \textcolor{blue}{20.20\%}) & 64.84 (\textcolor{blue}{$\uparrow$} \textcolor{blue}{14.93\%}) & 63.53 (\textcolor{blue}{$\uparrow$} \textcolor{blue}{13.62\%}) & 61.54 (\textcolor{blue}{$\uparrow$} \textcolor{blue}{11.63\%}) & 58.30 (\textcolor{blue}{$\uparrow$} \textcolor{blue}{8.39\%})  \\

\texttt{making scrambled eggs} & 71.61 (\textcolor{blue}{$\uparrow$} \textcolor{blue}{13.77\%}) & 79.90 (\textcolor{blue}{$\uparrow$} \textcolor{blue}{22.06\%}) & 82.66 (\textcolor{blue}{$\uparrow$} \textcolor{blue}{24.82\%}) & 80.72 (\textcolor{blue}{$\uparrow$} \textcolor{blue}{22.88\%}) & 78.63 (\textcolor{blue}{$\uparrow$} \textcolor{blue}{20.79\%})  \\

\texttt{paying with a credit card} & 74.52 (\textcolor{blue}{$\uparrow$} \textcolor{blue}{24.68\%}) & 75.53 (\textcolor{blue}{$\uparrow$} \textcolor{blue}{25.69\%}) & 73.12 (\textcolor{blue}{$\uparrow$} \textcolor{blue}{23.28\%}) & 77.32 (\textcolor{blue}{$\uparrow$} \textcolor{blue}{27.48\%}) & 65.85 (\textcolor{blue}{$\uparrow$} \textcolor{blue}{16.01\%})  \\

\rowcolor{maroon!12}\texttt{planting a tree} & 95.46 (\textcolor{blue}{$\uparrow$} \textcolor{blue}{35.18\%}) & 89.98 (\textcolor{blue}{$\uparrow$} \textcolor{blue}{29.70\%}) & 85.47 (\textcolor{blue}{$\uparrow$} \textcolor{blue}{25.19\%}) & 85.98 (\textcolor{blue}{$\uparrow$} \textcolor{blue}{25.70\%}) & 76.00 (\textcolor{blue}{$\uparrow$} \textcolor{blue}{15.72\%})  \\

\texttt{playing tennis} & 59.51 (\textcolor{blue}{$\uparrow$} \textcolor{blue}{9.33\%}) & 60.14 (\textcolor{blue}{$\uparrow$} \textcolor{blue}{9.96\%}) & 62.43 (\textcolor{blue}{$\uparrow$} \textcolor{blue}{12.25\%}) & 63.65 (\textcolor{blue}{$\uparrow$} \textcolor{blue}{13.47\%}) & 60.95 (\textcolor{blue}{$\uparrow$} \textcolor{blue}{10.77\%})  \\

\texttt{renovating a room} & 72.92 (\textcolor{blue}{$\uparrow$} \textcolor{blue}{21.86\%}) & 75.30 (\textcolor{blue}{$\uparrow$} \textcolor{blue}{24.24\%}) & 72.11 (\textcolor{blue}{$\uparrow$} \textcolor{blue}{21.05\%}) & 72.05 (\textcolor{blue}{$\uparrow$} \textcolor{blue}{20.99\%}) & 73.54 (\textcolor{blue}{$\uparrow$} \textcolor{blue}{22.48\%})  \\

\texttt{repairing a flat bicycle tire} & 80.63 (\textcolor{blue}{$\uparrow$} \textcolor{blue}{25.04\%}) & 83.22 (\textcolor{blue}{$\uparrow$} \textcolor{blue}{27.63\%}) & 80.89 (\textcolor{blue}{$\uparrow$} \textcolor{blue}{25.30\%}) & 82.12 (\textcolor{blue}{$\uparrow$} \textcolor{blue}{26.53\%}) & 77.30 (\textcolor{blue}{$\uparrow$} \textcolor{blue}{21.71\%})  \\

\rowcolor{customgreen!10}\texttt{riding on a bus} & 84.73 (\textcolor{blue}{$\uparrow$} \textcolor{blue}{29.75\%}) & 80.08 (\textcolor{blue}{$\uparrow$} \textcolor{blue}{25.10\%}) & 76.45 (\textcolor{blue}{$\uparrow$} \textcolor{blue}{21.47\%}) & 86.34 (\textcolor{blue}{$\uparrow$} \textcolor{blue}{31.36\%}) & 90.70 (\textcolor{blue}{$\uparrow$} \textcolor{blue}{35.72\%})  \\

\texttt{sending food back (in a restaurant)} & 70.47 (\textcolor{blue}{$\uparrow$} \textcolor{blue}{19.32\%}) & 73.10 (\textcolor{blue}{$\uparrow$} \textcolor{blue}{21.95\%}) & 71.93 (\textcolor{blue}{$\uparrow$} \textcolor{blue}{20.78\%}) & 64.22 (\textcolor{blue}{$\uparrow$} \textcolor{blue}{13.07\%}) & 66.61 (\textcolor{blue}{$\uparrow$} \textcolor{blue}{15.46\%})  \\

\texttt{sewing a button} & 76.73 (\textcolor{blue}{$\uparrow$} \textcolor{blue}{24.03\%}) & 81.05 (\textcolor{blue}{$\uparrow$} \textcolor{blue}{28.35\%}) & 80.59 (\textcolor{blue}{$\uparrow$} \textcolor{blue}{27.89\%}) & 77.56 (\textcolor{blue}{$\uparrow$} \textcolor{blue}{24.86\%}) & 76.23 (\textcolor{blue}{$\uparrow$} \textcolor{blue}{23.53\%})  \\

\texttt{taking a bath} & 77.71 (\textcolor{blue}{$\uparrow$} \textcolor{blue}{24.96\%}) & 85.67 (\textcolor{blue}{$\uparrow$} \textcolor{blue}{32.92\%}) & 81.47 (\textcolor{blue}{$\uparrow$} \textcolor{blue}{28.72\%}) & 81.65 (\textcolor{blue}{$\uparrow$} \textcolor{blue}{28.90\%}) & 77.57 (\textcolor{blue}{$\uparrow$} \textcolor{blue}{24.82\%})  \\

\texttt{taking a child to bed} & 68.52 (\textcolor{blue}{$\uparrow$} \textcolor{blue}{14.26\%}) & 74.16 (\textcolor{blue}{$\uparrow$} \textcolor{blue}{19.90\%}) & 69.31 (\textcolor{blue}{$\uparrow$} \textcolor{blue}{15.05\%}) & 70.35 (\textcolor{blue}{$\uparrow$} \textcolor{blue}{16.09\%}) & 68.11 (\textcolor{blue}{$\uparrow$} \textcolor{blue}{13.85\%})  \\

\texttt{taking a driving lesson} & 72.96 (\textcolor{blue}{$\uparrow$} \textcolor{blue}{18.67\%}) & 74.08 (\textcolor{blue}{$\uparrow$} \textcolor{blue}{19.79\%}) & 72.93 (\textcolor{blue}{$\uparrow$} \textcolor{blue}{18.64\%}) & 76.74 (\textcolor{blue}{$\uparrow$} \textcolor{blue}{22.45\%}) & 75.44 (\textcolor{blue}{$\uparrow$} \textcolor{blue}{21.15\%})  \\

\texttt{taking a shower} & 78.23 (\textcolor{blue}{$\uparrow$} \textcolor{blue}{22.62\%}) & 79.28 (\textcolor{blue}{$\uparrow$} \textcolor{blue}{23.67\%}) & 79.36 (\textcolor{blue}{$\uparrow$} \textcolor{blue}{23.75\%}) & 78.61 (\textcolor{blue}{$\uparrow$} \textcolor{blue}{23.00\%}) & 75.06 (\textcolor{blue}{$\uparrow$} \textcolor{blue}{19.45\%})  \\

\texttt{washing dishes} & 65.23 (\textcolor{blue}{$\uparrow$} \textcolor{blue}{14.99\%}) & 69.43 (\textcolor{blue}{$\uparrow$} \textcolor{blue}{19.19\%}) & 67.94 (\textcolor{blue}{$\uparrow$} \textcolor{blue}{17.70\%}) & 66.78 (\textcolor{blue}{$\uparrow$} \textcolor{blue}{16.54\%}) & 68.03 (\textcolor{blue}{$\uparrow$} \textcolor{blue}{17.79\%})  \\

\texttt{washing one s hair} & 76.49 (\textcolor{blue}{$\uparrow$} \textcolor{blue}{19.37\%}) & 84.70 (\textcolor{blue}{$\uparrow$} \textcolor{blue}{27.58\%}) & 82.76 (\textcolor{blue}{$\uparrow$} \textcolor{blue}{25.64\%}) & 78.70 (\textcolor{blue}{$\uparrow$} \textcolor{blue}{21.58\%}) & 79.09 (\textcolor{blue}{$\uparrow$} \textcolor{blue}{21.97\%})  \\
\bottomrule
\end{tabular}
}
\caption{
The table shows the performance of \texttt{Llama-3-8B} finetuned over 5 scenarios (presented in the columns). We observe a boost in performance (highlighted in blue) when compared to the MCQA-based evaluation. 
% In general, we observe generalization across similar scenarios.
% Success rate for \textbf{Llama-3(8B)} ,fine-tuned on source scenarios(columns) and validated on target scenarios(rows) on zero-shot task.
}
\label{tab:llama3-finetuned}
% \vspace{-4mm}
\end{table*}

\begin{figure*}
    \centering
    \includegraphics[width=\linewidth]{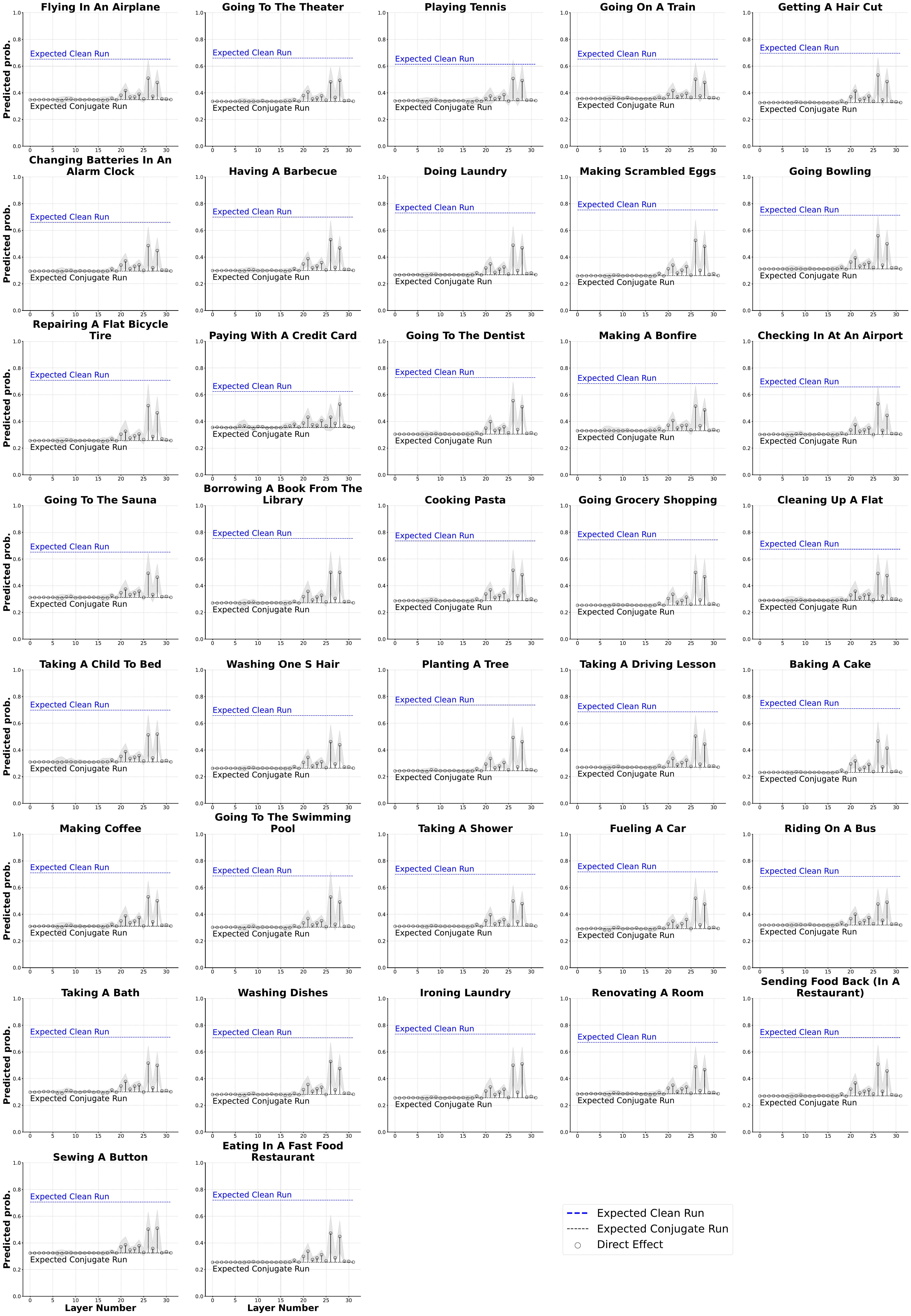}
     \caption{ The figure shows the direct effect of path patching from the clean run to the conjugate run, leading to deviations starting at layer 20 and reinforced by the following layers (maximum deviations observed at layer 26 and layer 28), highlighting the role of particular layers in commonsense reasoning.}
    \label{fig:decision-localization-all-scenarios}
\end{figure*}

\end{document}